\documentclass{article}

\usepackage[preprint]{neurips_2026}

\usepackage[utf8]{inputenc} 
\usepackage[T1]{fontenc}    
\usepackage{hyperref}       
\usepackage{url}            
\usepackage{booktabs}       
\usepackage{amsfonts}       
\usepackage{nicefrac}       
\usepackage{microtype}      
\usepackage{xcolor}         

\usepackage{amsmath}
\usepackage{amssymb}
\usepackage{mathtools}
\usepackage{amsthm}
\usepackage{tikz}
\usetikzlibrary{positioning}
\usepackage[capitalize,noabbrev]{cleveref}
\usepackage{graphicx}
\usepackage{subcaption}

\theoremstyle{plain}
\newtheorem{theorem}{Theorem}[section]
\newtheorem{proposition}[theorem]{Proposition}

\theoremstyle{definition}

\theoremstyle{remark}

\usepackage{algorithm}
\usepackage{algpseudocode}

\title{AETDICE: Unified Framework and Offline Optimization for Nonlinear Multi-Objective RL}

\author{
    Woosung Kim\textsuperscript{1*} \quad
    Youngjun Suh\textsuperscript{1*} \quad
    Jinho Lee\textsuperscript{1*} \quad
    Jongmin Lee\textsuperscript{2\dag} \quad
    Byung-Jun Lee\textsuperscript{1,3\dag} \\
    \\
    \textsuperscript{1}Korea University \quad
    \textsuperscript{2}Yonsei University \quad
    \textsuperscript{3}Gauss Labs Inc. \\
    \texttt{\{wsk208,jinho0997,youngjunsuh, byungjunlee\}@korea.ac.kr} \\
    \texttt{jongminlee@yonsei.ac.kr} \\
    \textsuperscript{*}Equal contribution \quad
    \textsuperscript{\dag}Corresponding authors
}

\begin{document}

\maketitle

\begin{abstract}
  Optimizing nonlinear preferences in multi-objective reinforcement learning (MORL) is essential for capturing complex trade-offs like risk aversion or fairness. However, such non-linearity has historically bifurcated nonlinear MORL objectives into two distinct paradigms: Scalarized Expected Return (SER) and Expected Scalarized Return (ESR). While SER requires global-level optimization and ESR requires non-Markovian policies, leading to fragmented optimization strategies, we bridge this divide through the Aggregation–Expectation–Transformation (AET) framework. By unifying both criteria through a tripartite decomposition of scalarization, AET provides a principled foundation for general nonlinear MORL. Building on this framework, we propose AETDICE, a tractable offline RL algorithm for AET objectives. By utilizing DICE-style density-ratio estimation in an augmented state space, AETDICE enables sample-based optimization from static datasets. Our framework resolves long-standing barriers and captures respective trade-offs induced by AET framework, which existing methods fail to address.
\end{abstract}

\section{Introduction}
Multi-objective reinforcement learning (MORL) optimizes sequential decisions under multiple, often competing objectives~\cite{van2014multi, cheung2019regret, yang2019generalized}. The trade-off among objectives is governed by a scalarization function that maps vector-valued returns to a scalar objective. Linear scalarization reduces to standard RL via a weighted sum, while nonlinear scalarization captures richer preferences such as diminishing returns, fairness and efficiency~\cite{roijers2013survey, agarwal2022multi}.



In nonlinear MORL, two canonical criteria arise from the ordering of scalarization and expectation: the Scalarized Expected Return (SER), which applies nonlinearity to the expected return, and the Expected Scalarized Return (ESR), which applies nonlinearity to each trajectory's return. SER and ESR represent different decision-making contexts with fundamentally different optimal policy structures, and existing methods have accordingly been developed separately for each~\cite{roijers2018multi, hayes2021practical, hayes2025expected}.


We focus on the offline setting, where policies must be learned from a fixed dataset. Offline methods exist for linear scalarization~\cite{zhu2023scaling, lin2024policy} and SER~\citep{kim2025fairdice}, but ESR remains unaddressed: trajectory-level nonlinearity makes the optimal action depend on previously accumulated rewards, and learning such reward-conditioned policies from fixed data has not been explored. Moreover, applying nonlinear utility to each trajectory's return vector yet aggregating the results linearly can neglect lower-performing utility dimensions—the same imbalance that nonlinear aggregation corrects in SER. While expectation-level nonlinearity can counteract this imbalance between nonlinear trajectory-level utilities, no existing method addresses this.

We address both gaps through the Aggregation–Expectation–Transformation (AET) framework, which decomposes nonlinear scalarization into a trajectory-level transformation ($F$), expectation, and an expectation-level aggregation ($G$). AET subsumes SER and ESR as special cases while enabling a richer class of objectives that combines both levels of nonlinearity. To optimize AET objectives offline, we introduce AETDICE—offline \emph{AET} optimization via \emph{DI}stribution \emph{C}orrection \emph{E}stimation. We propose a transformed reward that absorbs nonlinear $F$ into per-step rewards, enabling the first offline ESR optimization, and develop a finite-horizon DICE formulation over the augmented state space to handle the global aggregation of concave $G$. Empirically, AETDICE optimizes ESR, SER, and novel AET objectives within a single framework, revealing distinct policy behaviors that arise from the interaction of both levels of nonlinearity.

\section{Preliminaries}
\subsection{Finite-horizon MOMDPs and MORL objectives}
We consider a finite-horizon multi-objective Markov Decision Process (MOMDP) $\mathcal{M} = (\mathcal{S}, \mathcal{A}, P, \allowbreak H, \allowbreak p_0, \allowbreak \mathbf{r})$, where $\mathcal{S}$ and $\mathcal{A}$ denote the state and action spaces, $P(s' \mid s, a)$ is a time-homogeneous transition kernel, $H$ is the horizon, $p_0(s)$ is the initial-state distribution, and 
$\mathbf{r}(s,a) \in \mathbb{R}^m$ specifies an $m$-dimensional reward vector for $m$ objectives. A time-dependent policy $\pi(a|s,t)$ induces a trajectory distribution over $\tau = (s_0, a_0, \ldots, s_{H-1}, a_{H-1})$, and the cumulative multi-objective return is given by $\mathbf{R}(\tau) = \sum_{t=0}^{H-1} \mathbf{r}(s_t, a_t)$.

Multi-objective reinforcement learning (MORL) employs a scalarization function $u:\mathbb{R}^m \to \mathbb{R}$ to convert vector-valued returns into a scalar objective. The classical choice is linear scalarization, $u(\mathbf{x})=\mathbf{w}^{\top}\mathbf{x}$, which yields
\begin{equation}
    J_{\mathrm{lin}}(\pi; \mathbf{w}) = \mathbb{E}_{\tau \sim \pi}[\mathbf{w}^\top \mathbf{R}(\tau)],
\end{equation}
where $\mathbf{w}\in\mathbb{R}^m$ specifies linear trade-off between objectives. Since linear $u$ commutes with the expectation, the problem reduces to single-objective RL with scalar reward $r_{\mathbf{w}}=\mathbf{w}^{\top}\mathbf{r}$. To capture more general preference structures, nonlinear MORL allows $u$ to be nonlinear. In this setting, the expectation and scalarization no longer commute, giving rise to two canonical optimality criteria: the \emph{scalarized expected return} (SER) and the \emph{expected scalarized return} (ESR),
\begin{equation}
J_{\mathrm{SER}}(\pi;u)=u\left(\mathbb{E}_{\tau\sim\pi}[\mathbf{R}(\tau)]\right), \qquad J_{\mathrm{ESR}}(\pi;u)=\mathbb{E}_{\tau\sim\pi}[u(\mathbf{R}(\tau))].
\end{equation}
SER and ESR represent different decision-making contexts and place different roles on the scalarization $u$. Under SER, $u_{\mathrm{SER}}$ aggregates the expected return vector into a scalar objective and is typically concave, promoting balanced performance across objectives at the expectation level~\cite{kim2025fairdice, ruadulescu2020utility}.
Under ESR, $u_{\mathrm{ESR}}$ encodes a trajectory-level preference
and is not restricted to concave functions~\cite{cheung2019regret, fan2023welfare}. This distinction leads to fundamentally different optimal policy structures and optimization approaches, establishing the two central paradigms of nonlinear MORL.

\subsection{Optimization challenges in nonlinear MORL}
Figure~\ref{fig:momdp_example} illustrates how optimal policies differ across MORL with linear, SER, and ESR objectives in a simple two-step MOMDP with $u_{\mathrm{NSW}}(\mathbf{x})=\sum_{i=1}^m \log x_i$ (Nash Social
Welfare~\cite{kaneko1979nash}). Nonlinear MORL produces behaviors absent in linear MORL: ESR selects an action no linear MORL policy chooses, and SER yields a stochastic optimal policy. Beyond these behavioral differences, each criterion poses a distinct optimization challenge.

\begin{figure}
    \centering
    \begin{minipage}[c]{0.42\textwidth}
        \raggedright
        \begin{tikzpicture}[
            node distance=1.8cm,          
            >=stealth,
            auto,
            thick,
            state/.style={
                circle,
                draw=black,
                fill=white,
                minimum size=0.5cm,
                font=\small
            }
        ]
            \node[state] (s0) {$s_0$};
            \node[state] (s1) [right=of s0] {$s_1$};
            \node[state] (term) [right=of s1] {$s_T$};
            \draw[->] (s0) -- node {${a_0}:(0,0)$} (s1);
            \draw[->] (s1) edge[bend left=45] node {${a_1}:(9,1)$} (term);
            \draw[->] (s1) -- node {${a_2}:(4,4)$} (term);
            \draw[->] (s1) edge[bend right=45] node[swap] {${a_3}:(1,9)$} (term);
        \end{tikzpicture}
    \end{minipage}%
\begin{minipage}[c]{0.57\textwidth}
    \centering
    \footnotesize                         
    \setlength{\tabcolsep}{3pt}           
    \begin{tabular}{l|ccc}
        \toprule
        \textbf{Obj.} & \textbf{Optimal $\pi^{*}$} & $u_{\text{NSW}}(\mathbb{E}[\mathbf{R}])$ & $\mathbb{E}[u_{\text{NSW}}(\mathbf{R})]$ \\
        \midrule
        SER & $0.5\,a_1 + 0.5\,a_3$ & $\boldsymbol{\log(25)}$ & $\log(9)$ \\
        ESR & $a_2$ & $\log(16)$ & $\boldsymbol{\log(16)}$ \\
        Linear & $a_1$ or $a_3$ & $\log(9)$ & $\log(9)$ \\
        \bottomrule
    \end{tabular}
\end{minipage}
    \caption{Optimal policies of MORL with linear, SER, and ESR objectives in a two-step MOMDP with $u_{\text{NSW}}$. The optimal linear MORL policy selects the action corresponding to the higher-weighted objective. Full derivation in Appendix~\ref{app:example_ser_esr_linear}.}
    \label{fig:momdp_example}
\end{figure}



\paragraph{SER challenge: global-level optimization.}
In standard RL, the contribution of each state-action pair can be evaluated locally through rewards or advantages, enabling Bellman-style dynamic programming. In SER, however, nonlinear scalarization is applied outside the expectation, coupling all state-action pairs globally: the marginal contribution of any action depends on the policy's overall expected return vector, not on local information alone. This precludes local improvement and necessitates optimization at the level of the policy-induced distribution. In Figure~\ref{fig:momdp_example}, action probabilities must be optimized jointly, and the SER-optimal policy is stochastic—strictly outperforming all deterministic policies, unlike in single-objective RL or linear MORL. Prior work~\cite{kim2025fairdice} addresses this by reformulating the problem as convex optimization over occupancy measures, bypassing Bellman updates entirely.

\paragraph{ESR challenge: Non-Markovian policy.}
ESR applies nonlinearity at the trajectory level, making the optimal action depend not only on the current state but also on the accumulated reward $\mathbf{R}^{\mathrm{acc}}_t := \sum_{k=0}^{t-1} \mathbf{r}(s_k,a_k)$. For any $t>1$, the ESR objective can be written as
\begin{align*}
    J_{\mathrm{ESR}}(\pi;u_{\mathrm{ESR}})= \mathbb{E}_{\pi}\left[u_{\mathrm{ESR}}\left(\mathbf{R}_{t}^{\text{acc}}+\sum_{k=t}^{H-1}\mathbf{r}(s_k,a_k)\right)\right].
\end{align*}
Due to the nonlinearity, the future reward and $\mathbf{R}_{t}^{\text{acc}}$ cannot be optimized separately. The optimal action at state $s_t$ therefore depends on $\mathbf{R}_{t}^{\text{acc}}$: trajectories reaching the same state may require different actions depending on their reward history. ESR-optimal policies cannot, in general, be Markovian over the original state space~\cite{cheung2019regret, fan2023welfare}. Replacing $\mathbf{r}(s_0,a_0)$ with $=(2,0)$ in Figure~\ref{fig:momdp_example} shifts the ESR-optimal action at $s_1$ from $a_3$ to $a_2$, confirming that the optimal action depends on the reward history (derivation in Appendix~\ref{app:example_ser_esr_linear}). Prior works on ESR optimization~\cite{yu2023fair, siddiquelearning, peng2025multi} address this by augmenting the state with $\mathbf{R}^{\mathrm{acc}}_t$, restoring Markovianity in the augmented space.

\section{AET: A Unified Framework for Nonlinear Scalarization in MORL}
\label{sec:AET}
To address offline optimization of nonlinear trajectory-wise utilities and their nonlinear aggregation at the expectation level, we present AET, a unified framework that accommodates both levels of nonlinearity within a single formulation. AET decomposes nonlinear scalarization into three components: Aggregation (A), Expectation (E), and Transformation (T). The AET objective is
\begin{align}
\label{eq:AET-obj}
    J_{AET}(\pi;F,G)=G\!\left(
\mathbb{E}_{\tau \sim \pi}\!\left[F(\mathbf{R}(\tau))\right]
\right),
\end{align}
where $F : \mathbb{R}^m \to \mathbb{R}^n$ is a trajectory-level transformation composed of $n$ trajectory-wise utilities $F(\mathbf{R})=(f_1(\mathbf{R}), \allowbreak \dots, \allowbreak f_n(\mathbf{R}))$, and $G : \mathbb{R}^n \to \mathbb{R}$ is an expectation-level aggregation. We assume each $f_i$ is smooth and $G$ is concave to ensure a well-behaved optimization landscape.

Each $f_i$ encodes a distinct preference over the return vector, mapping the $m$-dimensional return vector to $n$ utility dimensions whose expectations are then aggregated by $G$. For instance, $f_i(\mathbf{R})=\log(R_i)$ captures diminishing marginal utility, amplifying gains in low-performing objectives; $f_i(\mathbf{R})=\exp(R_i)$ captures increasing marginal utility, amplifying gains in already high-performing objectives; and non-convex Cobb-Douglas $f_{i}(R_1,R_2)=R_1^{\rho}/(R_2)^{1-\rho}$ captures the efficiency trade-off ratio between two objectives. Linear $G$ aggregates the expected utilities linearly, while concave $G$ amplifies gains in lower-performing exptected utilities, encouraging balanced performance across all utility dimensions.

\paragraph{Landscape of Nonlinear MORL} 
The AET formulation organizes existing nonlinear MORL objectives by where nonlinearity is placed. Based on the nonlinearity of $F$ and $G$, we identify four regimes: (i) Linear MORL (linear $F$, linear $G$), recovered by $F(\mathbf{R})=\mathbf{R}$
and $G(\mathbf{v})=\mathbf{w}^{\top}\mathbf{v}$; (ii) SER-MORL (linear $F$, concave $G$), recovered by retaining the
identity transformation but adopting a concave aggregation
$G$; (iii) ESR-MORL (nonlinear $F$, linear $G$), recovered by setting $F = (f_1,\dots,f_n)$ with $G(\mathbf{v})=\mathbf{1}^{\top}\mathbf{v}$; and (iv) AET-MORL (nonlinear $F$, concave $G$), recovered by aggregating $F = (f_1,\dots,f_n)$ with concave $G$. We note that ESR-MORL can also be described as $F=f_1+\cdots+f_n$ with $G(v)=v$, reducing to a single scalar utility; we retain the vector representation to clearly distinguish it from AET-MORL, where concave $G$ preserves the distinction among multiple utility dimensions.

This taxonomy makes explicit which regimes lack offline methods: while Linear MORL and SER-MORL are addressed by prior work~\cite{zhu2023scaling,lin2024policy, kim2025fairdice}, no existing approach handles~\textbf{ESR-MORL} or~\textbf{AET-MORL}. ESR-MORL has been studied in online and value iteration setting~\cite{yu2023fair, siddiquelearning, peng2023nonlinear}, but (i) no offline method addresses the additional challenge of distribution shift, and (ii) current approaches cannot be extended to optimizing AET objectives with two nonlinearities, as they do not account for nonlinear aggregation $G$. The AET objective and its offline optimization are novel to this work. We fill these gaps by addressing each source of nonlinearity in turn: \cref{sec: ESR} tackles nonlinear $F$, \cref{sec: SER} tackles nonlinear $G$, and their combination yields AETDICE.

\section{Augmented MOMDP for Nonlinear Transformation F}
\label{sec: ESR}
We present an augmented MOMDP framework for optimizing nonlinear transformation $F$. While we adopt the augmented state space from prior work~\cite{peng2025multi} to restore Markovianity, we define a transformed reward that captures the incremental contribution of each action to $F$. This serves two purposes: it enables offline optimization of ESR-MORL by reducing it to standard single-objective RL in the augmented MOMDP, and it extends naturally to AET-MORL by reformulating it as SER-MORL in the augmented MOMDP.

\subsection{Transformed Reward in the Augmented Space}
We define the augmented finite-horizon MOMDP $\tilde{\mathcal{M}}=(\tilde{\mathcal{S}}, \mathcal{A}, \tilde{P}, H, \tilde{p}_0, \tilde{\mathbf{r}})$, where each augmented state is given by $\tilde{s}_t = (s_t, \mathbf{R}^{\mathrm{acc}}_t,t)$. The initial augmented state is $\tilde{s}_0 = (s_0, \mathbf{0}, 0)$ with initial distribution $\tilde{p}_0(\tilde{s}_0) = p_0(s_0)$. The transition kernel \(\tilde{P}\) is defined as
\begin{align}
    \begin{split}
        \tilde{P}(\tilde{s}_{t+1} | \tilde{s}_t, a_t) = P(s_{t+1} &| s_t, a_t) \cdot \mathbb{I}[\mathbf{R}^{\mathrm{acc}}_{t+1} = \mathbf{R}^{\mathrm{acc}}_t + \mathbf{r}(s_t,a_t)], \\
    \end{split}
\end{align}
where the timestep advances deterministically from $t$ to $t+1$. Since reward accumulation is absorbed into the transition dynamics and our objective focuses on the trajectory-wise 
utility $F(\mathbf{R}(\tau))$, we define a transformed reward to 
capture the incremental contribution of each action to $F$:
\begin{align}
\label{def:augmented_reward}
    \tilde{\mathbf{r}}(\tilde{s}_t,a_t) = F\big(\mathbf{R}^{\mathrm{acc}}_t+\mathbf{r}(s_t,a_t)\big) - F\big(\mathbf{R}^{\mathrm{acc}}_t\big), \quad \text{for} \ t\geq 1
\end{align}
with $\tilde{\mathbf{r}}(\tilde{s}_0,a_0)=F(\mathbf{r}(s_0,a_0))$. This induces an exact telescoping sum: $\sum_{t=0}^{H-1} \tilde{\mathbf{r}}(\tilde{s}_t,a_t) = F(\mathbf{R}(\tau))$, reducing the trajectory-level objective to a standard cumulative reward in the augmented MOMDP. When $F$ is linear, the transformed 
reward reduces to 
$\tilde{\mathbf{r}}(\tilde{s}_t,a_t) = F(\mathbf{r}(s_t,a_t))$, 
which no longer depends on $\mathbf{R}^{\mathrm{acc}}_t$; 
the optimal policy is then independent of the accumulated 
reward, making the augmentation redundant.

Under this construction, a policy must be conditioned on the augmented state $\tilde{s}_t$ to account for the dependence on accumulated rewards. We denote such policies as $\tilde{\pi}(a | \tilde{s}_t)$, which are Markovian in $\tilde{\mathcal{M}}$ but non-Markovian in the original MOMDP $\mathcal{M}$. This reduces ESR-MORL to solving single-objective RL in $\tilde{\mathcal{M}}$, as we show below.

\subsection{Offline ESR-MORL}
As described in~\cref{sec:AET}, when $G$ is linear, it commutes with the expectation, allowing us to collapse the AET objective into a single-utility case ($n$=1) by defining $\tilde{F} = G \circ F \colon \mathbb{R}^m \to \mathbb{R}$. ESR-MORL then becomes standard 
single-objective RL in $\tilde{\mathcal{M}}$, as the trajectory-level nonlinearity is fully absorbed into the transformed reward. This enables existing offline RL algorithms, such as IQL~\cite{kostrikov2021offline}, to be applied with only minor modifications to account for the augmented state $\tilde{s}_t$.

\begin{proposition}[Reduction to single-objective RL 
under linear $G$]
\label{prop:equivalence}
Let $G$ be linear. Then for any policy $\tilde{\pi}$ 
in the augmented MOMDP $\tilde{\mathcal{M}}$,
\begin{align}
    J_{\mathrm{AET}}(\tilde{\pi}; F, G) 
    = G\!\left(
\mathbb{E}_{\tau \sim \pi}\!\left[F(\mathbf{R}(\tau))\right]
\right)= \mathbb{E}_{\tau \sim \tilde{\pi}}\!\left[\tilde{F}(\mathbf{R(\tau)})\right] = \mathbb{E}_{\tau \sim \tilde{\pi}}\!\left[
    \sum_{t=0}^{H-1} \tilde{r}(\tilde{s}_t, a_t)\right],
\end{align}
where $\tilde{r}$ is defined in~\eqref{def:augmented_reward} 
with $\tilde{F} = G \circ F$.
\end{proposition}
\paragraph{Comparison with existing ESR methods.} Prior ESR optimization methods~\cite{yu2023fair, siddiquelearning} modify the value function update, rather than transforming the reward, to account for trajectory-wise nonlinearity, achieving effective online optimization. However, this approach does not align with two goals in our framework: offline optimization and AET optimization. Designing a value function update that simultaneously handles both nonlinearity and distribution shift is nontrivial. Our transformed reward decouples these concerns: the nonlinearity of $F$ is absorbed into the reward, while the value function update relies entirely on existing offline RL methods. Furthermore, value-function-based approaches are inherently local and incompatible with the global optimization required by concave $G$ in AET-MORL. This motivates the visitation-distribution-based formulation in the augmented MOMDP, which we develop in the next section. 

In addition, prior work on ESR~\cite{peng2025multi} conditions the policy on accumulated reward vector $\mathbf{R}^{\text{acc}}_t$ and assigns the transformed return $F(\mathbf{R}(\tau))$ only as a terminal reward. In contrast, our formulation provides a denser reward signal $\tilde{\mathbf{r}}$ that captures the incremental contribution of each action to $F(\mathbf{R}(\tau))$, making it amenable for sample-based optimization.

\section{Convex Optimization for Nonlinear Aggregation G}
\label{sec: SER}
With nonlinear $F$ absorbed into the transformed reward, optimizing AET-MORL reduces to SER-MORL in $\tilde{\mathcal{M}}$: maximizing a concave $G$ over multiple expected utilities. To handle the global-level optimization required by concave $G$, we extend the Distribution 
Correction Estimation (DICE) framework~\cite{kim2025fairdice, lee2021optidice} to the 
augmented MOMDP $\tilde{\mathcal{M}}$, yielding AETDICE—offline \emph{AET} optimization 
via \emph{Di}stribution \emph{C}orrection 
\emph{E}stimation.

\subsection{Visitation Distribution in the Augmented MOMDP}
In the augmented MOMDP, the visitation distribution induced by a policy $\tilde{\pi}$ is defined as $
d^{\tilde{\pi}}_t(\tilde{s},a)
\;:=\;
\Pr(\tilde{s}_t=\tilde{s}, a_t = a \mid \tilde{\pi})$,
representing the probability that $\tilde{\pi}$ visits the augmented state–action pair $(\tilde{s},a)$ at timestep $t$. While the timestep is already encoded within the augmented state $\tilde{s}$, we retain the index $t$ for notational clarity in the derivations below. Crucially, the expected utility vector becomes linear in the visitation distribution $d^{\tilde{\pi}}_t$:
\begin{equation}
\mathbb{E}_{\tau \sim \tilde{\pi}}\!\left[F(\mathbf{R(\tau)})\right] =\mathbb{E}_{\tau \sim \tilde{\pi}}\!\left[\sum_{t=0}^{H-1} \tilde{\mathbf{r}}(\tilde{s}_t,a_t)\right] = \sum_{t=0}^{H-1}\sum_{\tilde{s},a} d^{\tilde{\pi}}_t(\tilde{s},a)\,\tilde{\mathbf{r}}(\tilde{s},a).
\end{equation}
Therefore, AET-MORL reduces to a convex optimization 
problem over visitation distribution in 
$\tilde{\mathcal{M}}$, subject to finite-horizon Bellman 
flow constraints that ensure the validity of $\mathbf{d} = \{d_t\}_{t=0}^{H-1}$:
\begin{subequations}
\label{eq:aet_occ_form}
\begin{align}
\max_{\mathbf{d}\geq0} \quad & G\!\left(\sum_{t=0}^{H-1}
\sum_{\tilde{s},a}
d_t(\tilde{s},a)\,
\tilde{\mathbf{r}}(\tilde{s},a)
\right) \label{eq:aet_obj} \\
\text{s.t.} \quad & \sum_{a} d_0(\tilde{s},a)
= \tilde{p}_0(\tilde{s}), \quad \forall \tilde{s} 
\label{eq:bf_init} \\
& \sum_{a'} d_{t+1}(\tilde{s}',a')
= \sum_{\tilde{s},a}
d_t(\tilde{s},a)\,
\tilde{P}(\tilde{s}' \mid \tilde{s},a), \quad 
\forall \tilde{s}',\, t \in \{0, \dots, H-2\},
\label{eq:bf_flow}
\end{align}
\end{subequations}
where 
$\tilde{s}' = (s', \mathbf{R}^{\mathrm{acc}}_t + 
\mathbf{r}(s,a), t+1)$ denotes the next augmented state. This formulation enables the global optimization 
required by SER-like objectives with nonlinear $G$, 
and the optimal policy inducing optimal $\mathbf{d}^{*}$ can be recovered via $\tilde{\pi}_t^{*}(a|\tilde{s})
=d^{*}_t(\tilde{s},a)/\sum_{a}d^{*}_t(\tilde{s},a)$. However, solving this convex optimization directly requires access to the transition dynamics of $\tilde{\mathcal{M}}$. Moreover, even when the original state and action spaces are finite, the augmented state space $\tilde{\mathcal{S}}$ is generally continuous, as the accumulated reward $\mathbf{R}^{\mathrm{acc}}_t$ takes continuous values under general reward functions. This makes the convex formulation impractical even in tabular domains, motivating a sample-based approach.

\subsection{AETDICE: Regularized AET for sample-based optimization}

To derive a sample-based objective from \eqref{eq:aet_occ_form} in the offline setting, AETDICE extends the DICE-RL 
framework to 
the augmented MOMDP with two key modifications: (i) $\phi$-divergence regularization 
applied to the per-timestep data distribution $d_t^D$, correcting distribution 
shift at each step of the finite horizon, 
and (ii) a slack variable $\mathbf{k}$ that decouples 
the concave aggregation $G$ from the expectation 
over $\mathbf{d}$. The regularized primal problem is:
\begin{align}
\max_{\mathbf{d} \in \mathcal{D}, \mathbf{k} \in \mathbb{R}^m}  G(\mathbf{k}) - \beta \sum_{t=0}^{H-1} D_{\phi}(d_t \| d^D_t), \quad \text{s.t.} \quad \sum_{t=0}^{H-1} \sum_{\tilde{s}, a} d_t(\tilde{s}, a) \tilde{\mathbf{r}}(\tilde{s}, a) = \mathbf{k}, \label{eq:aetdice_primal_main}
\end{align}
where $\mathcal{D}$ denotes the set of non-negative visitation distribution satisfying constraints~\eqref{eq:bf_init}--\eqref{eq:bf_flow}, $\beta > 0$ is a regularization coefficient, and $D_{\phi}(d_t\,\|\,d^D_t) :=\sum_{\tilde s, a} d_t^D(\tilde{s}, a) \phi \left( d_t(\tilde s, a)/d_t^D(\tilde s, a) \right)$ is the $\phi$-divergence between $d_t$ and the data distribution $d_t^D$ at time step $t$. The full derivation, required assumptions, and complete algorithm are provided in Appendix~\ref{app:aetdice_dual}.

\paragraph{Dual formulation.}
We form the Lagrangian of~\eqref{eq:aetdice_primal_main}
by introducing $\boldsymbol{\mu} \in \mathbb{R}^n$ for 
the slack constraint, and 
time-indexed multipliers $\nu_t(\tilde{s})$ for the 
Bellman flow 
constraints~\eqref{eq:bf_init}--\eqref{eq:bf_flow}, replacing the time-independent multiplier used in standard DICE approaches~\citep{kim2025fairdice, lee2021optidice} to reflect the 
finite-horizon structure of the augmented MOMDP. The primal problem becomes 
$\max_{\mathbf{d}\geq0,\mathbf{k}} 
\min_{\boldsymbol{\nu},\boldsymbol{\mu}} 
\mathcal{L}$, where:
\begin{align}\mathcal{L}(\mathbf{d}, \mathbf{k}, \boldsymbol{\nu}, \boldsymbol{\mu}):= G(\mathbf{k}) - \boldsymbol{\mu}^\top \mathbf{k}+ \sum_{\tilde{s}} \nu_0(\tilde{s}) \tilde{p}_0(\tilde{s}) - \beta \sum_{t=0}^{H-1} D_{\phi}(d_t || d^D_t) + \sum_{\tilde{s}, a,t} d_t(\tilde{s}, a) e_{\boldsymbol{\nu},\boldsymbol{\mu},t}(\tilde{s}, a)\nonumber \end{align}
with $e_{\boldsymbol{\nu},\boldsymbol{\mu},t}(\tilde{s}, a) 
= \boldsymbol{\mu}^\top \tilde{\mathbf{r}}(\tilde{s}, a) 
+ \mathbb{E}_{\tilde{s}'}[\nu_{t+1}(\tilde{s}')] 
- \nu_t(\tilde{s})$ and $\nu_H(\tilde{s}) = 0$ for 
consistency.
\paragraph{Derivation of training objectives.} Under strong duality, we swap the optimization order to $\min_{\boldsymbol{\nu},\boldsymbol{\mu}} \max_{\mathbf{d} \ge 0,\mathbf{k}} \mathcal{L}$. The inner maximization over $\mathbf{d}$ yields a closed-form optimal visitation distribution ratio $w^*_t(\tilde{s}, a) = d^*_t(\tilde{s}, a) / d^D_t(\tilde{s}, a)=\max\left(0,(\phi')^{-1} \left(e_{\boldsymbol{\nu},\boldsymbol{\mu},t}(\tilde{s}, a)/\beta \right)\right)$. We leverage Fenchel conjugates $\phi_+^*(y) := \sup_{x\geq0} 
\{ xy - \phi(x) \}$ and 
$G^*(\mathbf{y}) := \sup_{\mathbf{x}} 
\{ \mathbf{x}^{\top}\mathbf{y} + G(\mathbf{x}) \}$ to simplify the min--max optimization and yield an unconstrained objective that depends only on expectations over $\tilde{p}_0$ and the offline 
data $d^D_t$:
\begin{align}
\label{eq:aetdice_dual_final}
\min_{\boldsymbol{\nu}, \boldsymbol{\mu}} \mathcal{L}_{AET}(\boldsymbol{\nu}, \boldsymbol{\mu}):= \mathbb{E}_{\tilde{p}_0} [\nu_0(\tilde{s})] + \sum_{t=0}^{H-1} \mathbb{E}_{d_t^D} \left[ \beta \phi_+^* \left( e_{\boldsymbol{\nu},\boldsymbol{\mu},t}(\tilde{s}, a)/\beta\right) \right] +G^{*}(-\boldsymbol{\mu}).\end{align}
After minimizing 
$\mathcal{L}_{\mathrm{AET}}$ to obtain $\nu^*$ and 
$\boldsymbol{\mu}^*$, the policy that 
induces the optimal visitation distribution can be extracted using 
the optimal visitation ratio $w^*_t$ via weighted behavior cloning following~\cite{lee2021optidice}:
\begin{align*}
\max_{\pi}\; \mathbb{E}_{(\tilde{s},a) \sim d_t^D}
\!\left[w^{*}_{t}(\tilde{s},a) 
\log \pi(a|\tilde{s}_t)\right].
\end{align*}

\paragraph{Comparison with FairDICE.}
AETDICE differs from the DICE framework~\cite{kim2025fairdice, lee2021optidice} in three key aspects: (i) prior DICE approaches impose Bellman flow constraints in a discounted infinite-horizon MDP, whereas AETDICE imposes finite-horizon Bellman flow constraints over the augmented state space~\eqref{eq:bf_init}--\eqref{eq:bf_flow}; (ii) the Lagrangian multipliers and policy are conditioned on the augmented state $\tilde{s}$, and offline data $d_t^D$ must be preprocessed to include the accumulated reward vector $\mathbf{R}^{\mathrm{acc}}_t$; and (iii) AETDICE extends FairDICE, which handles only SER objectives, to support a broader class of nonlinear objectives including ESR and AET, though it operates in a higher-dimensional state space. 

\section{Experiments}
\label{sec:experiments}
We conduct two sets of experiments to demonstrate that AETDICE enables offline optimization of AET objectives with nonlinear $F$ and $G$. First, we focus on offline Fair MORL with concave $F$ and $G$, optimizing and comparing different fairness objectives within a single framework. Second, we go beyond concavity to optimize convex and non-convex trajectory-wise utilities, along with their concave aggregation, on complex continuous-control domains.

\paragraph{Algorithms.}
We evaluate offline methods corresponding to each nonlinear MORL regime identified in \cref{sec:AET}:
(1)~\textbf{OptiDICE}~\citep{lee2021optidice}: a DICE-based offline RL method with linear $F$ and linear $G$;
(2)~\textbf{FairDICE}~\citep{kim2025fairdice}: a DICE-based offline MORL method with linear $F$ and concave $G$;
(3)~\textbf{ESR-IQL} (ours): an IQL-based method on the augmented MOMDP with nonlinear $F$ and linear $G$; and
(4)~\textbf{AETDICE-\textit{Obj}} (ours): a DICE-based method on the augmented MOMDP, applicable to any combination of $F$ and $G$.

\subsection{Offline Fair MORL: ESR, SER, and BSR}
\begin{figure}[t]
    \centering
    \captionsetup{font=footnotesize}
    \begin{subfigure}[b]{0.28\linewidth}
        \centering
        \includegraphics[
        width=\linewidth]{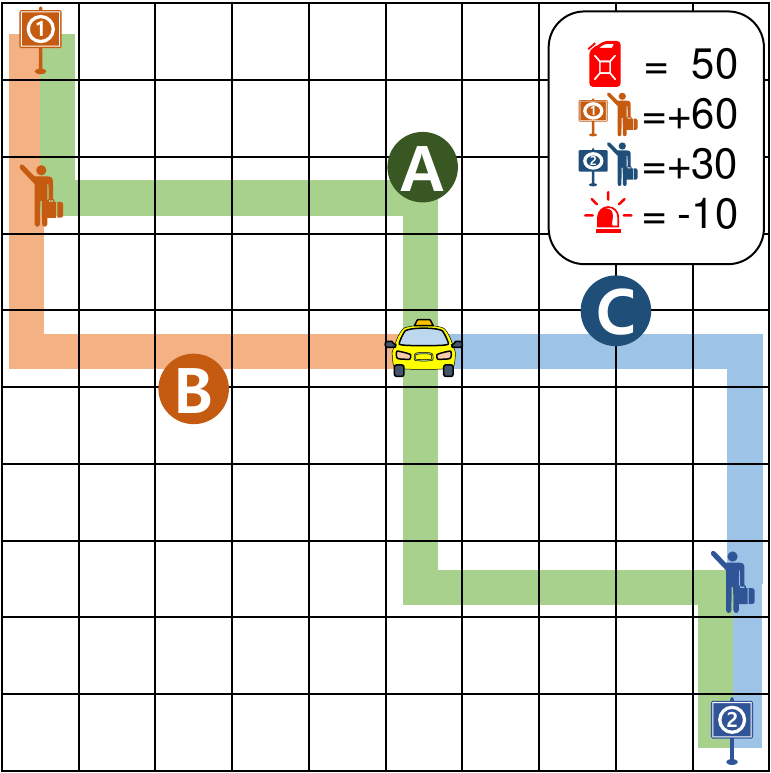}
    \end{subfigure}
    \hfill
    \begin{subfigure}[b]{0.70\linewidth}
        \centering
        \includegraphics[trim={2pt 2pt 2pt 2pt}, clip,
        width=\linewidth]{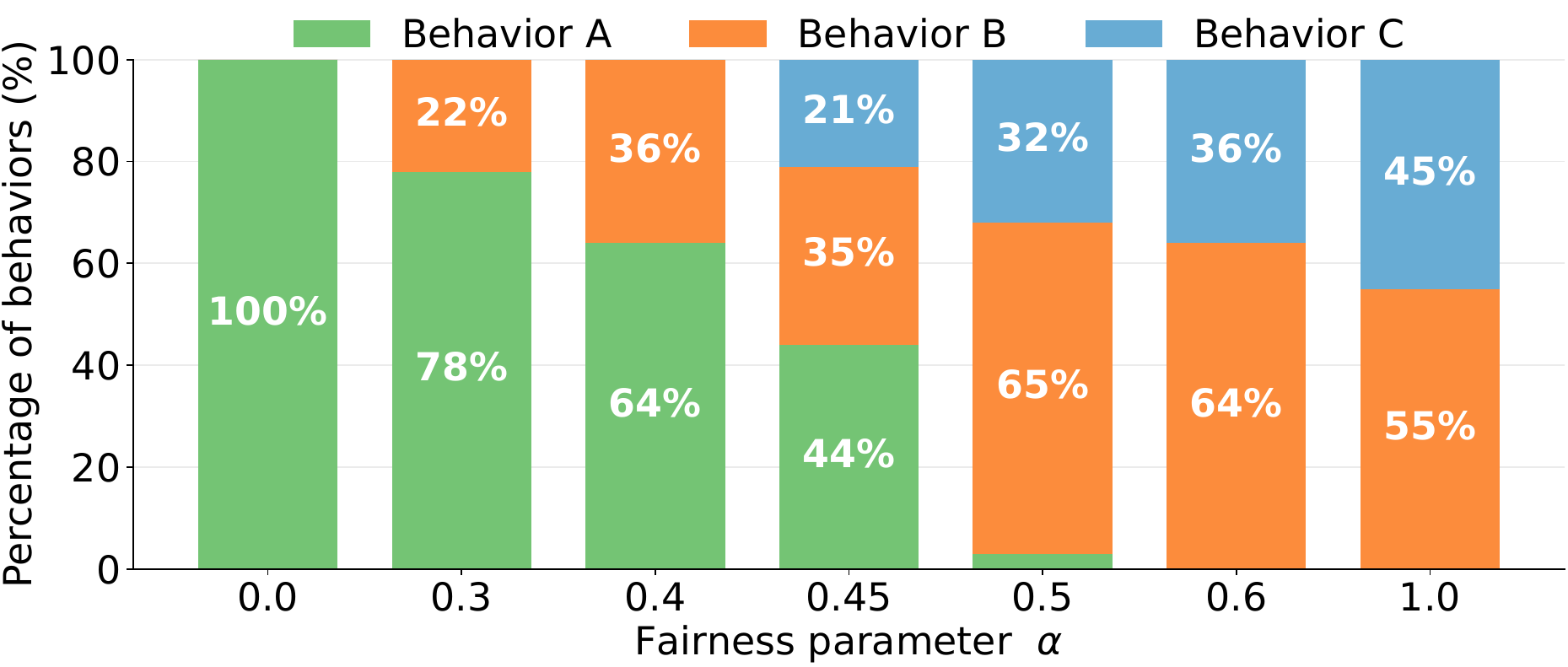} 
    \end{subfigure}
    \caption{\textbf{(a)}~Fair-Taxi: a taxi agent serving two passenger groups. Each objective corresponds to serving a passenger group. \textbf{(b)}~Distribution of behavior patterns across varying nonlinear MORL objectives. Full details in Appendix~\ref{app:env_details}.}
    \label{fig:environments}
    \vskip -0.18in
\end{figure}

Fair MORL~\cite{kim2025fairdice, park2024max} applies concave scalarization to promote equitable performance across objectives, where improvements to lower-performing objectives yield larger gains in overall utility. As discussed in \cref{sec:AET}, the AET framework subsumes existing Fair MORL criteria as special cases. To further bridge trajectory-level and expectation-level fairness, we adopt the $\alpha$-fairness utility~\citep{mo2002fair} to construct \textbf{Balanced Scalarized Return (BSR)}, a novel objective family parameterized by $\alpha$ that interpolates between trajectory-level and expectation-level fairness:
\begin{equation}
J_{\mathrm{BSR}}(\pi;\alpha) = G_{\alpha}\!\left(
\mathbb{E}_{\tau\sim\pi}\!\left[F_{\alpha}(\mathbf{R}(\tau))
\right]\right), \qquad 
F_{\alpha}(\mathbf{x}) = u_{1-\alpha}(\mathbf{x}), 
\quad 
G_{\alpha}(\mathbf{x}) = \sum_{i=1}^{n} u_{\alpha}(x_i).
\label{eq:bsr}
\end{equation}
where $u_{\alpha}(x) = \frac{x^{1-\alpha}-1}{1-\alpha}$ for $\alpha \neq 1$ and $u_{1}(x) = \log x$. By varying $\alpha \in [0, 1]$, BSR controls the 
relative nonlinearity applied at the trajectory level 
($F_{\alpha}$) and the aggregation level ($G_{\alpha}$), 
recovering ESR at $\alpha = 0$ and SER at $\alpha = 1$. We analyze ESR-, BSR-, and SER-optimal policies obtained by AETDICE and illustrate their distinct behaviors in Fair-Taxi environment (\cref{fig:environments}). Details and full results including other environments including MO-PointMaze-3obj are presented in Appendix~\ref{app:env_details}.

\paragraph{Behavioral differences.}
Figure~\ref{fig:environments}(a) illustrates three episode-level behavior patterns of optimal nonlinear MORL policies in Fair-Taxi: (A)~serving both groups, (B)~serving only group~1, and (C)~serving only group~2. Figure~\ref{fig:environments}(b) shows their distribution over 100 episodes. The ESR-optimal policy consistently produces pattern~(A), balancing return across objectives within each episode. The SER-optimal policy yields a stochastic mixture of (B) and (C), where each trajectory specializes in a single objective and balance is achieved only in expectation. The BSR-optimal policy smoothly interpolates between these extremes: as $\alpha$ increases toward SER, specialized trajectories gradually emerge, while lower $\alpha$ recovers the balanced ESR behavior.

\begin{figure}[h]
  \vskip -0.0in
  \begin{center}
    \centerline{\includegraphics[width=0.8\columnwidth]{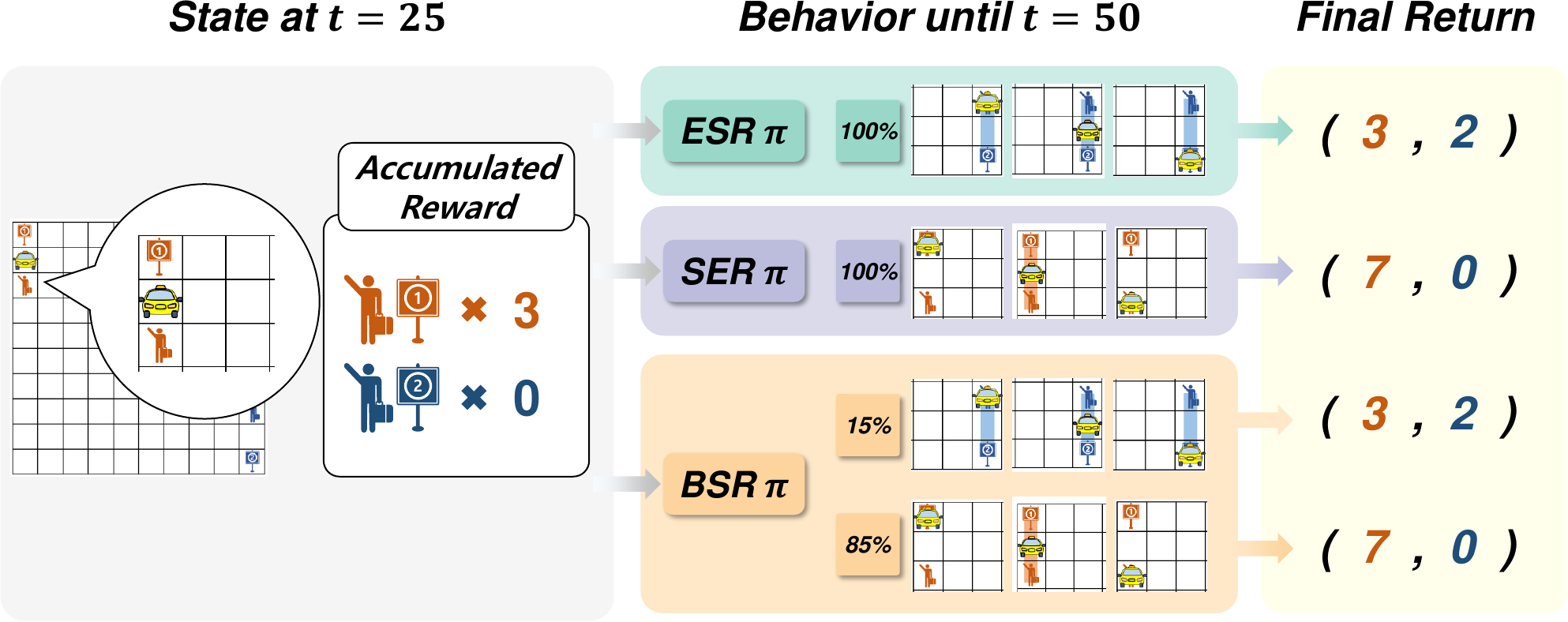}}
    \caption{
        Non-Markovian behavior of ESR and BSR-optimal policies in Fair-Taxi environment.
    }
    \label{nonmarkovian}
  \end{center}
  \vskip -0.3in
\end{figure}

\paragraph{Non-Markovian policy.}
The non-Markovian structure formalized in \cref{sec: ESR} manifests concretely in the ESR- and BSR-optimal policies. Under these objectives, the concavity of $F$ transforms the original reward through $\mathbf{R}^{\mathrm{acc}}_t$: as one objective accumulates higher returns, its transformed marginal reward diminishes, steering the policy toward the under-served objective. Since the transformed reward depends on $\mathbf{R}^{\mathrm{acc}}_t$, the optimal action from the same state varies with trajectory history, making the policy inherently non-Markovian. In contrast, the SER-optimal policy is unaffected, as its reward does not depend on $\mathbf{R}^{\mathrm{acc}}_t$. Figure~\ref{nonmarkovian} captures a state at $t=25$ where group~1 has already been served three times. The ESR-optimal policy shifts to group~2 due to diminished marginal reward for group~1, while the SER-optimal policy continues serving group~1 independently of history. The BSR-optimal policy ($\alpha=0.45$) stochastically chooses between the two strategies.

\begin{figure}[ht]
  \vskip 0.0in
  \begin{center}
    \centerline{\includegraphics[width=\columnwidth]{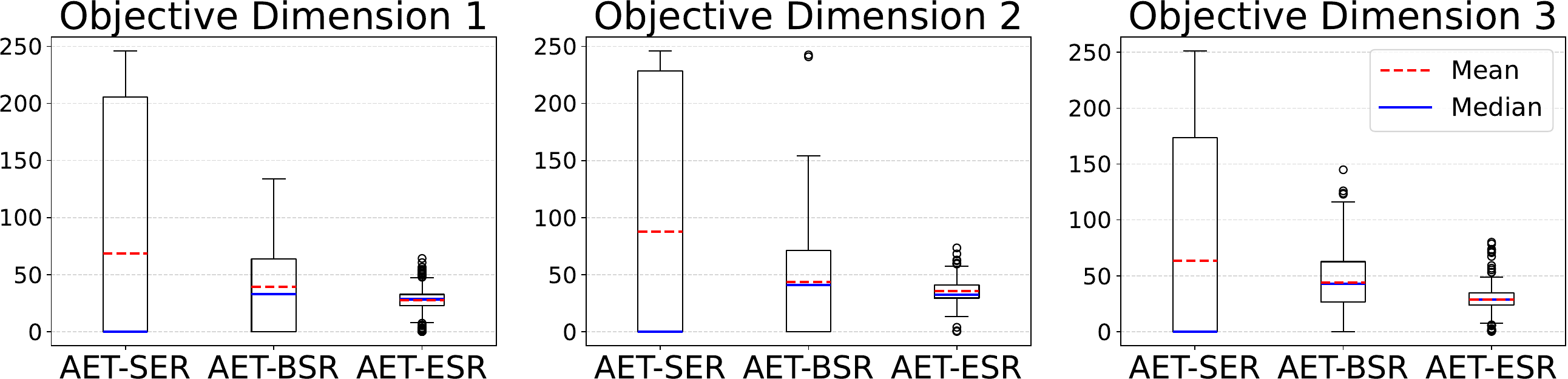}}
    \caption{
    Quantile distributions of per-trajectory returns in MO-PointMaze-3obj environment
    }
    \label{quantile_plot_Taxi}
  \end{center}
  \vskip -0.25in
\end{figure}

\paragraph{Quantitative evaluation.} The quantile plot in Figure~\ref{quantile_plot_Taxi} illustrates how different Fair MORL objectives shape the per-trajectory return distribution across individual objective dimensions. The SER-optimal policy achieves higher expected returns but with large variance across trajectories, while the ESR-optimal policy yields lower expected returns with significantly reduced variance. The BSR-optimal policy lies between these extremes, offering a tunable trade-off. Prior work has noted that SER is suited for repeated policy execution and ESR for single-shot deployment~\cite{roijers2013survey}. BSR accommodates practitioners who seek the benefits of repeated execution while constraining the variability across objectives, interpolating between the two regimes via $\alpha$.

\subsection{Beyond Concavity: General AET Objectives}
Beyond Fair MORL, the AET framework accommodates diverse trajectory-level utilities and their aggregation by varying $F$ and $G$. We demonstrate this on the D4MORL benchmark~\cite{zhu2023scaling}, a more complex continuous-control domain, with four configurations: (i) convex $F(\mathbf{x}) = \exp(\mathbf{x}/50)$ with linear $G(\mathbf{x})$, which prioritizes high-performing objectives; (ii) the same convex $F$ with concave $G(\mathbf{x}) = \sum_i \log(x_i)$, which combines trajectory-level specialization with expectation-level balance; and (iii) non-convex Cobb-Douglas $F(\mathbf{x}) = x_1^{\rho}/{x_2}^{1-\rho}$ with linear $G$, a single trajectory-wise utility case that optimizes trajectory-wise efficiency; and (iv) non-convex Cobb-Douglas $f_1$ for efficiency and concave safety-threshold $f_2(x)=\log(x-R_{th})$ designed to penalize collapse, aggregated by concave $G$.

\begin{figure}
  \centering
  \begin{subfigure}{0.33\linewidth}
    \includegraphics[width=\linewidth]{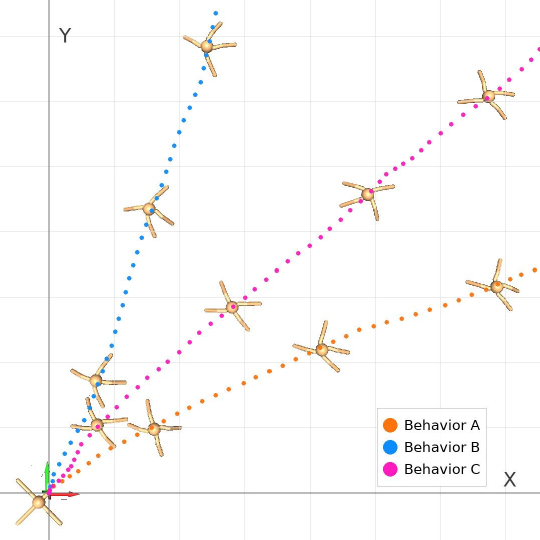}
  \end{subfigure}\hspace{0.1em}%
  \begin{subfigure}{0.33\linewidth}
    \includegraphics[width=\linewidth]{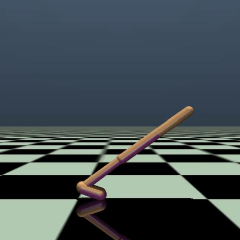}
  \end{subfigure}\hspace{0.1em}%
  \begin{subfigure}{0.33\linewidth}
    \includegraphics[width=\linewidth]{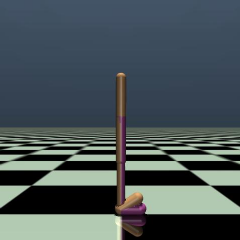}
  \end{subfigure}
  \caption{\textbf{(a)}~MO-Ant-v2: AETDICE with convex $F$ and concave $G$ produces a stochastic mixture of specialized behaviors (Behavior~A: 67\%, Behavior~B: 33\%), while Linear MORL with equal preference weights produces balanced diagonal movement (Behavior~C). \textbf{(b)}~Over-optimization of Cobb-Douglas utility leads to agent collapse. \textbf{(c)}~Adding safety utility $f_2$ maintains Cobb-Douglas efficiency while preventing collapse.}
  \label{fig:behaviors}
\end{figure}

\paragraph{Convex utility F.}
We apply convex $F$ with increasing marginal utility to MO-Ant, where the two objectives correspond to movement in orthogonal directions. Convex $F$ amplifies gains from concentrating on a single objective, encouraging the policy to commit to one direction per episode rather than distributing effort across both. Under Linear MORL, the agent moves diagonally according to the preference weights (Behavior~C in Figure~\ref{fig:behaviors}). With convex $F$ and linear $G$~(i), however, the amplification effect causes the policy to specialize exclusively in one direction (Behavior~A), even under equal weighting.
Pairing the same convex $F$ with concave $G$~(ii) addresses this at the expectation level: the concave aggregation amplifies under-served utility dimensions, producing a stochastic mixture of Behavior~A and Behavior~B, where each trajectory specializes in a different direction yet balance is maintained across episodes—a behavior that is difficult to achieve with linear $G$, as it requires precise weight tuning to counteract the amplification of convex $F$.
\paragraph{Cobb-Douglas utility.}
We focus on MO-Walker2d and MO-Halfcheetah as two objectives, forward locomotion and energy consumption, present a natural efficiency trade-off. The Cobb-Douglas utility $F(\mathbf{x}) = x_1^{\rho}/(x_2)^{1-\rho}$ is convex in one objective and concave in the other, making it non-convex overall. Despite this, the utility has a natural interpretation in these environments: by treating energy consumption as the denominator, the Cobb-Douglas utility effectively optimizes the movement-to-energy ratio, capturing trajectory-wise efficiency. By varying $\rho$, one controls the relative emphasis on forward locomotion versus energy consumption. However, at excessively low values (e.g., $\rho=0.1$), the MO-Walker2d agent frequently collapses, as the utility overly prioritizes energy saving at the expense of maintaining stable locomotion. 

\paragraph{Cobb-Douglas utility with safety utility.}
To counteract the collapse caused by over-optimizing efficiency, we introduce a safety utility $f_2(x) = \log(x - R_{\mathrm{th}})$ that penalizes trajectories approaching a safety threshold, analogous to a log-barrier. This is aggregated with the Cobb-Douglas utility via concave $G$, forming a two-utility AET objective:
\begin{align}
    J_{AET}(\tilde{\pi})=\log\left(\mathbb{E}_{\tau\sim\tilde{\pi}}\left[\frac{R(\tau)^{\rho}}{C(\tau)^{1-\rho}}\right]\right)+\log\left(\mathbb{E}_{\tau\sim\tilde{\pi}}[R_s(\tau) - R_{th}]\right)
\end{align}
where $R_{th}=0.8H$ is set to 80\% of the maximum achievable safety return. Figure~\ref{fig:behaviors}(b,c) compares AETDICE policies optimizing Cobb-Douglas utility alone versus with the additional safety utility, showing that the safety utility successfully prevents the Walker2d agent from collapsing. These results demonstrate that the AET framework extends beyond concave transformations to handle a broader class of practical objectives. Full results are presented in Appendix~\ref{app:additional_exp}.

\begin{table}[h]
  \caption{Results on MO-Ant with convex trajectory-level utility $F$. Linear is abbreviated as Lin.}
  \label{tab:mo_ant}
  \begin{center}
    \begin{small}
      \begin{sc}
        \begin{tabular}{l|c|c}
          \toprule
          Method& convex $F$ + linear $G$ & convex $F$ + concave $G$ \\
          \midrule
          
            Lin-$[1.0, 0.0]$& $724.85 \pm 1.84$ & $12.89 \pm 0.02$\\
            Lin-$[0.5, 0.5]$& $684.53 \pm 7.53$ & $13.04 \pm 0.02$\\
            Lin-$[0.0, 1.0]$& $1221.13 \pm 28.26$ & $12.25 \pm 0.03$\\
          FairDICE& $681.95 \pm 24.28$ & $13.04 \pm 0.07$\\
          \midrule
          ESR-IQL& $\mathbf{1251.46 \pm 17.88}$ & $12.24 \pm 0.01$\\
          AET-ESR& $ 1067.22 \pm 6.81$ & $ 13.04\pm 0.12$\\
          AETDICE& $ 953.53 \pm 16.31$ & $\mathbf{13.69 \pm 0.05}$\\
          \bottomrule
        \end{tabular}
      \end{sc}
    \end{small}
  \end{center}
\end{table}


\section{Conclusion}
We presented AET, a unified framework that organizes nonlinear MORL objectives by the position of nonlinearity—trajectory-level transformation $F$ and expectation-level aggregation $G$—and identifies two previously unaddressed regimes: offline ESR-MORL and AET-MORL. To fill these gaps, we proposed AETDICE, which introduces a transformed reward that absorbs nonlinear $F$ into per-step rewards in an augmented MOMDP, and a finite-horizon DICE formulation that handles the global optimization required by concave $G$. Empirically, AETDICE optimizes ESR, SER, BSR, and novel AET objectives within a single framework, revealing distinct policy behaviors—including non-Markovian strategies and stochastic mixtures—that emerge from the interaction of both levels of nonlinearity. Limitations and broader impact are discussed in Appendix~\ref{app:limitations}.

\section{Acknowledgements}
This work was partly supported by Institute of Information $\&$ Communications Technology Planning $\&$ Evaluation (IITP) grant funded by the Korea government (MSIT) (No. RS-2022-II220311, Development of Goal-Oriented Reinforcement Learning Techniques for Contact-Rich Robotic Manipulation of Everyday Objects, No. RS-2024- 00457882, AI Research Hub Project, No. RS-2019- II190079, Artificial Intelligence Graduate School Program (Korea University), the IITP (Institute of Information $\&$ Communications Technology Planning $\&$ Evaluation)-ITRC (Information Technology Research Center) grant funded by the Korea government (Ministry of Science and ICT) (IITP-2026-RS-2024-00436857), BK21 Four project of the National Research Foundation of Korea, the National Research Foundation of Korea (NRF) grant funded by the Korea government (MSIT) (RS2025-00560367), the IITP under the Artificial Intelligence Star Fellowship support program to nurture the best talents (IITP-2026-RS-2025-02304828) grant funded by the Korea government (MSIT).
This work was also supported by the Institute of Information $\&$ Communications Technology Planning $\&$ Evaluation (IITP) grant (RS-2020-II201361, Artificial Intelligence Graduate School Program (Yonsei University), and the AI Computing Infrastructure Enhancement (GPU Rental Support) User Support Program funded by the Ministry of Science and ICT (MSIT) (No.~RQT-25-090109), Republic of Korea.

\bibliographystyle{unsrt}
\bibliography{references}


\newpage
\appendix
\section{Related Works}
\label{app:related_work}

\paragraph{Multi-Objective Reinforcement Learning}
Multi-objective reinforcement learning (MORL) studies sequential decision-making problems with vector-valued rewards, where no single policy is universally optimal under the Pareto dominance relation~\cite{roijers2013survey}. As a result, MORL methods often aim to compute a set of trade-off solutions (e.g., Pareto-optimal or coverage sets) so that a decision maker can later select a policy once preferences are specified~\cite{van2013scalarized, van2014multi}. A widely used instantiation is linear scalarization, which reduces the problem to a standard single-objective MDP by optimizing a weighted sum of objectives and thereby enables the use of conventional RL algorithms~\cite{lizotte2010efficient, van2013hypervolume}. This perspective has led to two main solution paradigms: set-based approaches, which learn a collection of Pareto-optimal policies, and preference-conditioned methods, which adapt a single policy to different trade-offs~\cite{roijers2013survey, van2013scalarized, van2014multi}.

\paragraph{Nonlinear Preference in MORL}
However, linear scalarization cannot capture richer preference structures such as fairness, balance, or risk sensitivity~\cite{skalse2023limitations}. To address this limitation, recent work has explored nonlinear scalarization based on concave utility functions applied to return vectors. Under appropriate smoothness and concavity conditions, such nonlinear objectives remain tractable to optimize~\cite{agarwal2022multi, peng2025multi}. This line of work includes welfare-oriented criteria such as Nash social welfare and Gini-based measures, which encourage equitable performance across objectives~\cite{fan2023welfare, siddique2023fairness}. Related approaches also study max–min formulations that explicitly promote balanced outcomes, often combined with entropy regularization to stabilize learning in model-free online settings~\cite{park2024max}. Together, these works reflect a broader shift from linear trade-offs toward more expressive preference modeling in MORL.

\paragraph{Offline RL and Offline MORL}
Offline reinforcement learning (RL) focuses on learning policies from fixed datasets without additional environment interaction, where distributional shift is a central challenge. Existing approaches address this issue through conservative value estimation ~\cite{kumar2020conservative}, implicit value regularization~\cite{kostrikov2021offline}, divergence-regularized optimization~\cite{lee2021optidice}, and return-conditioned sequence modeling~\cite{chen2021decision}.

In the multi-objective setting, most offline approaches rely on linear scalarization and require explicit preference conditioning during training or evaluation~\cite{wu2021offline, lin2024policy}. FairDICE~\cite{kim2025fairdice} adapts stationary distribution correction to a nonlinear welfare objective, showing that certain nonlinear criteria can be handled in offline MORL. However, it is tailored to a specific welfare formulation and does not directly generalize to broader classes of nonlinear objectives with different placements of nonlinearity in the return structure.

\section{Optimal Policies in the Two-Step MOMDP Example}
\label{app:example_ser_esr_linear}
\begin{figure}[H]
    \centering
    \begin{minipage}[c]{0.42\textwidth}
        \raggedright
        \begin{tikzpicture}[
            node distance=1.8cm,          
            >=stealth,
            auto,
            thick,
            state/.style={
                circle,
                draw=black,
                fill=white,
                minimum size=0.5cm,
                font=\small
            }
        ]
            \node[state] (s0) {$s_0$};
            \node[state] (s1) [right=of s0] {$s_1$};
            \node[state] (term) [right=of s1] {$s_T$};
            \draw[->] (s0) -- node {${a_0}:(0,0)$} (s1);
            \draw[->] (s1) edge[bend left=45] node {${a_1}:(9,1)$} (term);
            \draw[->] (s1) -- node {${a_2}:(4,4)$} (term);
            \draw[->] (s1) edge[bend right=45] node[swap] {${a_3}:(1,9)$} (term);
        \end{tikzpicture}
    \end{minipage}%
\begin{minipage}[c]{0.57\textwidth}
    \centering
    \footnotesize                         
    \setlength{\tabcolsep}{3pt}           
    \begin{tabular}{l|ccc}
        \toprule
        \textbf{Obj.} & \textbf{Optimal $\pi^{*}$} & $u_{\text{NSW}}(\mathbb{E}[\mathbf{R}])$ & $\mathbb{E}[u_{\text{NSW}}(\mathbf{R})]$ \\
        \midrule
        SER & $0.5\,a_1 + 0.5\,a_3$ & $\boldsymbol{\log(25)}$ & $\log(9)$ \\
        ESR & $a_2$ & $\log(16)$ & $\boldsymbol{\log(16)}$ \\
        Linear & $a_1$ or $a_3$ & $\log(9)$ & $\log(9)$ \\
        \bottomrule
    \end{tabular}
\end{minipage}
    \caption{Optimal policies of MORL with linear, SER, and ESR objectives in a two-step MOMDP with $u_{\text{NSW}}(\mathbf{x})=\sum_{i}\log(R_i)$.}
    \label{fig:momdp_example_app}
\end{figure}
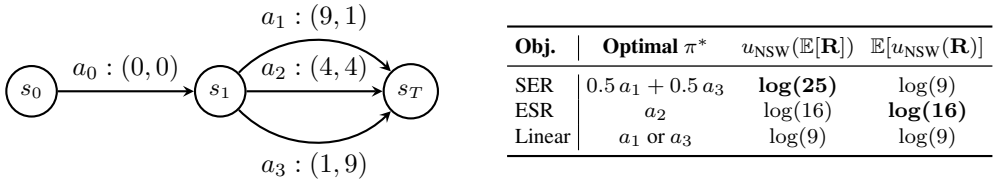

In this section, we derive the optimal policies for linear, SER, and ESR objectives in the two-step MOMDP shown in Figure~\ref{fig:momdp_example_app} (reproduced from Figure~\ref{fig:momdp_example} for convenience). Through these derivations, we illustrate the distinct optimization challenges that arise from nonlinear MORL objectives.

\paragraph{Linear scalarization.}
For linear preference weights $[w_1, w_2] \in \mathbb{R}^2_{+}$, the linear MORL objective is defined as $J_{\mathrm{lin}}(\pi;\mathbf{w}) = \mathbb{E}_\pi[w_1 R_1 + w_2 R_2]$. The optimal policy selects $a_1$ when $w_1 > w_2$ and $a_3$ when $w_1 < w_2$, with both SER $u_{\text{NSW}}(\mathbb{E}[\mathbf{R}])$ and ESR $\mathbb{E}[u_{\text{NSW}}(\mathbf{R})]$ remaining constant at $\log 9$ in either case. The exception is $w_1 = w_2$, where $a_1$ and $a_3$ are equally optimal, making any stochastic mixture also optimal. However, Scalarized Expected Return (SER) is no longer constant in this case, as it varies with the mixing probability.

\paragraph{Scalarized Expected Return (SER).}
SER maximizes $J_{\mathrm{SER}}(\pi) = u_{\mathrm{NSW}}(\mathbb{E}_\pi[\mathbf{R}])$. Unlike the linear case, optimizing SER is not straightforward: letting $p_1$, $p_2$, $p_3$ denote the probabilities of selecting $a_1$, $a_2$, $a_3$, one must solve a convex optimization over the full policy distribution. This reflects a key challenge of SER optimization—the nonlinearity applies to the expected return, requiring global optimization over the entire policy rather than local, per-action evaluation. The convex optimization can be written as:
\begin{align}
\max_{p_1, p_2, p_3} \quad & \log(9p_1 + 4p_2 + p_3) + \log(p_1 + 4p_2 + 9p_3) \nonumber \\
\text{s.t.} \quad & p_1 + p_2 + p_3 = 1, \quad p_1, p_2, p_3 \geq 0.
\end{align}
where the optimal solution is $p_1^* = 0.5$, $p_2^* = 0$, $p_3^* = 0.5$, yielding a stochastic policy that mixes $a_1$ and $a_3$ equally.

\paragraph{Expected Scalarized Return (ESR).}
ESR maximizes $J_{\mathrm{ESR}}(\pi) = \mathbb{E}_\pi[u_{\mathrm{NSW}}(\mathbf{R})]$. Since the nonlinearity is applies before the expectation, each action can be evaluated independently: $u_{\mathrm{NSW}}(\mathbf{R}(a_1)) = \log 9$, $u_{\mathrm{NSW}}(\mathbf{R}(a_2)) = \log 16$, $u_{\mathrm{NSW}}(\mathbf{R}(a_3)) = \log 9$. The optimal policy is therefore deterministic: $\pi^{*}=a_2$.

\subsection{Demonstrating ESR non-Markovianity}
By changing the initial state-action pair $(s_0, a_0)$ such that the reward at $s_0$ becomes $(2, 0)$ instead of $(0, 0)$, the previously accumulated reward at $s_1$ shifts from $(0, 0)$ to $(2, 0)$. This changes each trajectory's return vector to $\mathbf{R}(a_1) = (11, 1)$, $\mathbf{R}(a_2) = (6, 4)$, $\mathbf{R}(a_3) = (3, 9)$, yielding $u_{\mathrm{NSW}}(\mathbf{R}(a_1)) = \log 11$, $u_{\mathrm{NSW}}(\mathbf{R}(a_2)) = \log 24$, $_{\mathrm{NSW}}(\mathbf{R}(a_3)) = \log 27$. The optimal action at $s_1$ shifts from $a_2$ to $a_3$, demonstrating that the ESR-optimal policy depends on the accumulated reward and is therefore non-Markovian in the original state space.

\section{Derivation of the AETDICE Loss Functions }
\label{app:aetdice_dual}
In this section, we provide the full derivation of the AETDICE loss functions. AETDICE assumes the augmented MOMDP $\tilde{\mathcal{M}}$ defined in \cref{sec: ESR} with the transformed reward:
\begin{align}
\tilde{\mathbf{r}}(\tilde{s}_t, a_t) = F\!\left(\mathbf{R}^{\mathrm{acc}}_t + \mathbf{r}(s_t, a_t)\right) - F\!\left(\mathbf{R}^{\mathrm{acc}}_t\right) \;\in\; \mathbb{R}^n.
\end{align}
The AET-MORL objective $\max_{\tilde{\pi}} J_{\mathrm{AET}}(\tilde{\pi};F,G)$ is reformulated to a convex optimization in terms of $\mathbf{d}=\{d_t(\tilde{s},a)\}_{t=0}^{H-1}$, the visitation distributions in $\tilde{\mathcal{M}}$ as
\begin{subequations}
\label{app-eq:aet_occ_form}
\begin{align}
\max_{\mathbf{d}\geq0} \quad & G\!\left(\sum_{t=0}^{H-1}
\sum_{\tilde{s},a}
d_t(\tilde{s},a)\,
\tilde{\mathbf{r}}(\tilde{s},a)
\right)  \\ 
\text{s.t.} \quad & \sum_{a} d_0(\tilde{s},a)
= \tilde{p}_0(\tilde{s}), \quad \forall \tilde{s} 
 \\ 
& \sum_{a'} d_{t+1}(\tilde{s}',a')
= \sum_{\tilde{s},a}
d_t(\tilde{s},a)\,
\tilde{P}(\tilde{s}' \mid \tilde{s},a), \quad 
\forall \tilde{s}',\, t \in \{0, \dots, H-2\},
\end{align}
\end{subequations}
where 
$\tilde{s}' = (s', \mathbf{R}^{\mathrm{acc}}_t + 
\mathbf{r}(s,a), t+1)$. No flow constraint is imposed at $t=H-1$, as all transitions from $d_{H-1}$ lead to the terminal state.
 
\subsection{Regularized Primal Problem}
\label{app:aetdice_primal}
To enable sample-based optimization of the problem in offline setting, we follow the derivation from~\cite{kim2025fairdice} by (i) adding $\phi$-divergence regularization against per-timestep data distribution $d^{D}_t$ and (ii) introducing a slack variable $\mathbf{k}\in\mathbb{R}^n$ to relocate the expected utility vector outside the concave aggregation: 
\begin{subequations}
\label{eq:aetdice_primal_app}
\begin{align}
\max_{\mathbf{d}\geq 0,\,\mathbf{k}\in\mathbb{R}^n}\quad
& G(\mathbf{k}) - \beta\sum_{t=0}^{H-1} D_\phi(d_t \,\|\, d_t^D) \label{eq:aet_obj_app}\\
\text{s.t.}\quad
& \sum_{t=0}^{H-1}\sum_{\tilde{s},a} d_t(\tilde{s},a)\,\tilde{\mathbf{r}}(\tilde{s},a) = \mathbf{k}, \label{eq:slack_app}\\
& \sum_a d_0(\tilde{s},a) = \tilde{p}_0(\tilde{s}),
  \qquad \forall \tilde{s}\in\tilde{\mathcal{S}}, \label{eq:bf0_app}\\
& \sum_{a'} d_{t+1}(\tilde{s}',a')
= \sum_{\tilde{s},a}
d_t(\tilde{s},a)\,
\tilde{P}(\tilde{s}' \mid \tilde{s},a), \quad 
\forall \tilde{s}',\, t \in \{0, \dots, H-2\},\label{eq:bft_app}
\end{align}
\end{subequations}
where no flow constraint is imposed at $t=H$, as all transitions from $d_{H-1}$ lead to the terminal state.

\paragraph{Assumptions.}
We require: (i) $G$ is concave and continuous; (ii) $\phi$ is strictly convex, with $\phi(1) = 0$; and (iii) the data distribution satisfies $d^D_t(\tilde{s},a) > 0$ for all reachable $(\tilde{s},a)$ and $t$, so that the $\phi$-divergence is well-defined and Slater's condition is satisfied. Assumption~(iii) is stronger than the corresponding coverage assumption in DICE-based offline RL~\cite{lee2021optidice}, which only requires coverage over the original state-action space. In the augmented MOMDP, coverage must additionally hold over the accumulated reward $\mathbf{R}^{\mathrm{acc}}_t$, yet the set of $\mathbf{R}^{\mathrm{acc}}_t$ values reachable from a given state is generally a strict subset of $\mathbb{R}^m$. This can be partially mitigated by initializing trajectories in the data with nonzero accumulated rewards, but the coverage requirement remains fundamentally more demanding than in the original MOMDP. 
 
\paragraph{Role of the slack variable k.}
The slack variable $\mathbf{k}$ is a technique first introduced in~\cite{kim2025fairdice}, which decouples the expectation over $\mathbf{d}$ from the concave aggregation $G$. By enforcing \eqref{eq:slack_app}, the expected transformed reward is computed separately and passed to $G$ through $\mathbf{k}$. This decoupling is essential for deriving an unbiased sample-based approximation of the Lagrangian dual: without $\mathbf{k}$, estimating $G\!\left(\sum_t \sum_{\tilde{s},a} d_t(\tilde{s},a)\,\tilde{\mathbf{r}}(\tilde{s},a)\right)$ from samples would require applying $G$ to an empirical average, which is biased by Jensen's inequality.

\subsection{Lagrangian Dual}
\label{app:aetdice_lagrangian}
We derive the Lagrangian dual of \eqref{eq:aetdice_primal_app} by introducing multipliers $\nu_t(\tilde{s})$ for each Bellman-flow constraints \eqref{eq:bf0_app}--\eqref{eq:bft_app}, and $\boldsymbol{\mu}\in\mathbb{R}^n$ for the slack constraint \eqref{eq:slack_app}:
\begin{align}
\max_{\mathbf{d}\geq 0,\mathbf{k}}\,\min_{\boldsymbol{\nu},\boldsymbol{\mu}} \ \mathcal{L}(\mathbf{d},\mathbf{k},\boldsymbol{\nu},\boldsymbol{\mu})
&= G(\mathbf{k})
  - \boldsymbol{\mu}^{\!\top}\!\left(\sum_{t,\tilde{s},a} d_t(\tilde{s},a)\,\tilde{\mathbf{r}}(\tilde{s},a) - \mathbf{k}\right)
  - \beta\sum_{t=0}^{H-1} D_\phi(d_t\,\|\,d_t^D) \nonumber\\
&\quad + \sum_{\tilde{s}} \nu_0(\tilde{s})\!\left(\tilde{p}_0(\tilde{s}) - \sum_a d_0(\tilde{s},a)\right) \nonumber\\
&\quad + \sum_{t=0}^{H-2}\sum_{\tilde{s}} \nu_t(\tilde{s}')\!\left(\sum_{\tilde{s},a} \tilde{P}(\tilde{s}'\mid\tilde{s},a)\,d_{t}(\tilde{s},a) - \sum_a d_{t+1}(\tilde{s}',a)\right). \label{eq:lagrangian_raw}
\end{align}
 
To enable sample-based evaluation of the dual function, we reorganize the last term of the dual function into expectations with respect to the visitation distribution $\mathbf{d}$:
\begin{align}
&\sum_{\tilde{s}} \nu_0(\tilde{s})\,\tilde{p}_0(\tilde{s})
  \;-\; \sum_{\tilde{s},a}\nu_0(\tilde{s})\,d_0(\tilde{s},a) \nonumber\\
&\quad +\; \sum_{\tilde{s},a} d_0(\tilde{s},a)\sum_{\tilde{s}'} \tilde{P}(\tilde{s}'\mid\tilde{s},a)\,\nu_1(\tilde{s}')
  \;-\; \sum_{\tilde{s},a}\nu_1(\tilde{s})\,d_1(\tilde{s},a) \nonumber\\
&\quad +\; \sum_{\tilde{s},a} d_1(\tilde{s},a)\sum_{\tilde{s}'} \tilde{P}(\tilde{s}'\mid\tilde{s},a)\,\nu_2(\tilde{s}') \;-\; \cdots \nonumber\\
&\quad -\; \sum_{\tilde{s},a} \nu_{H-1}(\tilde{s})\,d_{H-1}(\tilde{s},a) +\; \sum_{\tilde{s},a} d_{H-1}(\tilde{s},a)\sum_{\tilde{s}'} \tilde{P}(\tilde{s}'\mid\tilde{s},a)\,\underbrace{\nu_H(\tilde{s}')}_{:=\,0}.
\end{align}
 where we adopt $\nu_H(\tilde{s}) := 0$ for all $\tilde{s}\in\tilde{\mathcal{S}}$ for notational uniformity. This enables the derivation of the reorganized Lagrangian dual:
\begin{align} \mathcal{L}(d, \mathbf{k}, \boldsymbol{\nu}, \boldsymbol{\mu})= & G(\mathbf{k}) - \boldsymbol{\mu}^\top \mathbf{k} - \beta \sum_{t=0}^{H-1} D_{\phi}(d_t || d^D_t)+ \sum_{\tilde{s}} \nu_0(\tilde{s}) \tilde{p}_0(\tilde{s}) + \sum_{\tilde{s}, a,t} d_t(\tilde{s}, a) e_{\boldsymbol{\nu},\boldsymbol{\mu},t}(\tilde{s}, a) \nonumber\end{align}
where $e_{\boldsymbol{\nu},\boldsymbol{\mu},t}(\tilde{s}, a) 
= \boldsymbol{\mu}^\top \tilde{\mathbf{r}}(\tilde{s}, a) 
+ \mathbb{E}_{\tilde{s}'}[\nu_{t+1}(\tilde{s}')] 
- \nu_t(\tilde{s})$.
 
\subsection{Switching the order of optimization via strong duality}
\label{app:aetdice_swap}
We leverage strong duality of~\eqref{eq:aetdice_primal_app} enabled by the data coverage assumption and satisfaction of Slater's condition to swap the optimization order to $\min_{\boldsymbol{\nu},\boldsymbol{\mu}}\;\max_{\mathbf{d}\geq 0,\,\mathbf{k}}\;\mathcal{L}$. This yields a closed-form solution for $\mathbf{d}$ and enables simplification of the dual function via conjugate functions.

We define the per-timestep visitation ratio $w_t(\tilde{s},a) := \frac{d_t(\tilde{s},a)}{d_t^D(\tilde{s},a)}$ prior to obtaining the closed-form solution to further simplify the Lagrangian:
\begin{align}
\mathcal{L}
= G(\mathbf{k}) - \boldsymbol{\mu}^{\!\top}\mathbf{k}
  + \mathbb{E}_{\tilde{p}_0}[\nu_0(\tilde{s})]
  + \sum_{t=0}^{H-1} \mathbb{E}_{(\tilde{s},a)\sim d_t^D}\!\Big[ w_t(\tilde{s},a)\, e_{\boldsymbol{\nu},\boldsymbol{\mu},t}(\tilde{s},a) - \beta\,\phi(w_t(\tilde{s},a)) \Big].
\label{eq:L_with_w}
\end{align}

Given $(\boldsymbol{\nu},\boldsymbol{\mu},\mathbf{k})$, the inner maximization in \eqref{eq:L_with_w} decouples pointwise across $(t,\tilde{s},a)$:
\begin{align}
\max_{w\geq 0}\;\Big\{\, w\cdot e_{\boldsymbol{\nu},\boldsymbol{\mu},t}(\tilde{s},a) - \beta\,\phi(w) \,\Big\}.
\end{align}
The first-order condition $e_{\boldsymbol{\nu},\boldsymbol{\mu},t}(\tilde{s},a) - \beta\,\phi'(w) = 0$ yields the closed-form maximizer
\begin{align}
w_t^{*}(\tilde{s},a)
= \max\!\left(0,\;(\phi')^{-1}\!\left(\frac{e_{\boldsymbol{\nu},\boldsymbol{\mu},t}(\tilde{s},a)}{\beta}\right)\right),
\label{eq:wstar}
\end{align}
where the $\max(0,\cdot)$ enforces the non-negativity constraint.
 
\paragraph{Fenchel-conjugate substitution.}
Define the non-negative-restricted Fenchel conjugate of $\phi$,
\begin{align}
\phi^{*}_{+}(y) := \sup_{w\geq 0}\,\big\{\, w y - \phi(w) \,\big\}.
\end{align}
By a direct change of variables,
\begin{align}
\sup_{w\geq 0}\,\big\{\, w z - \beta\phi(w) \,\big\}
= \beta\,\sup_{w\geq 0}\,\Big\{\, w\,\tfrac{z}{\beta} - \phi(w) \,\Big\}
= \beta\,\phi^{*}_{+}\!\left(\tfrac{z}{\beta}\right).
\label{eq:phi_conj_substitution}
\end{align}
Substituting \eqref{eq:phi_conj_substitution} with $z = e_{\boldsymbol{\nu},\boldsymbol{\mu},t}(\tilde{s},a)$ into \eqref{eq:L_with_w} eliminates the inner maximization over $w_t$:
\begin{align}
\max_{\mathbf{d}\geq 0}\;\mathcal{L}
= G(\mathbf{k}) - \boldsymbol{\mu}^{\!\top}\mathbf{k}
  + \mathbb{E}_{\tilde{p}_0}[\nu_0(\tilde{s})]
  + \sum_{t=0}^{H-1} \mathbb{E}_{d_t^D}\!\left[ \beta\,\phi^{*}_{+}\!\left(\frac{e_{\boldsymbol{\nu},\boldsymbol{\mu},t}(\tilde{s},a)}{\beta}\right) \right].
\end{align} 
The remaining $\mathbf{k}$-dependent part of the Lagrangian is $G(\mathbf{k}) - \boldsymbol{\mu}^{\!\top}\mathbf{k}$, maximized unconstrained over $\mathbf{k}\in\mathbb{R}^n$. Following the convention used elsewhere in the paper, define the (concave) conjugate of $G$ as $G^{*}(\mathbf{y}) := \sup_{\mathbf{x}}\,\big\{\, \mathbf{x}^{\!\top}\mathbf{y} + G(\mathbf{x}) \,\big\}$.
Then
\begin{align}
\max_{\mathbf{k}\in\mathbb{R}^n}\,\big\{\, G(\mathbf{k}) - \boldsymbol{\mu}^{\!\top}\mathbf{k} \,\big\}
= \max_{\mathbf{k}\in\mathbb{R}^n}\,\big\{\, \mathbf{k}^{\!\top}(-\boldsymbol{\mu}) + G(\mathbf{k}) \,\big\}
= G^{*}(-\boldsymbol{\mu}).
\end{align}

Combining the two conjugate formulations reduces the dual problem to a sample-based loss function given by,
\begin{align}
\min_{\boldsymbol{\nu},\boldsymbol{\mu}}\;\mathcal{L}_{\mathrm{AET}}(\boldsymbol{\nu},\boldsymbol{\mu})
:= \mathbb{E}_{\tilde{p}_0}[\nu_0(\tilde{s})]
  + \sum_{t=0}^{H-1} \mathbb{E}_{(\tilde{s},a)\sim d_t^D}\!\left[ \beta\,\phi^{*}_{+}\!\left(\frac{e_{\boldsymbol{\nu},\boldsymbol{\mu},t}(\tilde{s},a)}{\beta}\right) \right]
  + G^{*}(-\boldsymbol{\mu}).
\label{eq:aetdice_final_dual}
\end{align}
Crucially, $\mathcal{L}_{\mathrm{AET}}$ is expressed entirely as expectations over $\tilde{p}_0$ and the offline dataset $\{d_t^D\}_{t=0}^{H-1}$. It does not require knowledge of the augmented transition kernel $\tilde{P}$ (the $\tilde{P}$-expectation inside $e_{\boldsymbol{\nu},\boldsymbol{\mu},t}$ is replaced by single-sample plug-in $\nu_{t+1}(\tilde{s}_{t+1})$ at offline transitions $(\tilde{s}_t,a_t,\tilde{s}_{t+1})$), nor does it require explicit access to the primal visitation distributions $\{d_t\}$. It can therefore be optimized using standard stochastic gradient descent on neural-network parametrizations $\nu_{\theta,t}$ and $\boldsymbol{\mu}$.
 
\subsection{Policy Extraction}
\label{app:aetdice_policy}
 
After minimizing \eqref{eq:aetdice_final_dual} to obtain $(\boldsymbol{\nu}^{*},\boldsymbol{\mu}^{*})$, the optimal visitation ratio $w_t^{*}$ is recovered from \eqref{eq:wstar} using $e_{\boldsymbol{\nu}^{*},\boldsymbol{\mu}^{*},t}$. The optimal augmented-state policy $\tilde{\pi}^{*}_t(a\mid\tilde{s})$ inducing the optimal visitation $d_t^{*}=w_t^{*}\,d_t^D$ is then extracted. In the tabular case, this reduces to direct normalization: $\tilde{\pi}^{*}_t(a\mid\tilde{s})
= \frac{w_t^{*}(\tilde{s},a)\,d_t^D(\tilde{s},a)}{\sum_{a'} w_t^{*}(\tilde{s},a')\,d_t^D(\tilde{s},a')}$. In continuous case, we use weighted behavior cloning following \cite{lee2021optidice}:
\begin{align}
\max_{\tilde{\pi}}\;\mathbb{E}_{(\tilde{s},a)\sim d_t^D}\!\left[ w_t^{*}(\tilde{s},a)\,\log \tilde{\pi}(a\mid\tilde{s}) \right].
\end{align}

\subsection{Choice of $\phi$-Divergence and Aggregation $G$}
\label{app:aetdice_concrete}

We use the $\chi^{2}$-divergence for $\phi$ and a piecewise-log function for the concave aggregation $G$.

\paragraph{$\chi^{2}$ divergence.}
$\phi(x) = \tfrac{1}{2}(x-1)^{2}$ for $x \ge 0$. Its convex conjugate is
\begin{align}
\phi^{*}_{+}(y) \;=\; \tfrac{1}{2}\bigl[\,[y+1]_{+}\,\bigr]^{2} \;-\; \tfrac{1}{2},
\qquad
w_{t}^{*}(\tilde{s},a) \;=\; \bigl[\,e_{\boldsymbol{\nu},\boldsymbol{\mu},t}(\tilde{s},a)/\beta + 1\,\bigr]_{+},
\end{align}
where $[\,\cdot\,]_{+} := \max(0,\cdot)$.

\paragraph{Piecewise-log aggregation.}
In practice, we avoid using $G(\mathbf{x})=\sum_{i}\log(x_i)$ directly, as logarithm is undefined for negative values that may arise during optimization. Instead, we use a piecewise extension $G(\mathbf{x}) = \sum_{i=1}^{d} g(x_{i})$ with
\begin{align}
g(x) \;=\;
\begin{cases}
\log(x), & x \ge 1, \\
-\tfrac{1}{2}(x-2)^{2} + \tfrac{1}{2}, & x < 1.
\end{cases}
\end{align}
which is concave, strictly increasing, and well-defined on all of $\mathbb{R}$. With $G^{*}(\boldsymbol{p}) := \sup_{\mathbf{x}}\{\boldsymbol{p}^{\!\top}\mathbf{x} + G(\mathbf{x})\}$, its conjugate is $G^{*}(-\boldsymbol{\mu}) = \sum_{i=1}^{d} g^{*}(-\mu_{i})$ with
\begin{align}
g^{*}(-\mu) \;=\;
\begin{cases}
-1 - \log(\mu), & 0 < \mu < 1, \\
\tfrac{1}{2}\mu^{2} - 2\mu + \tfrac{1}{2}, & \mu \ge 1.
\end{cases}
\end{align}

\subsection{Full Algorithm of AETDICE}
We present the full algorithm of AETDICE in Algorithm~\ref{alg:aetdice}. A key implementation consideration is that the Lagrangian multipliers $\nu_t(\tilde{s})$ are defined per timestep, and the neural network must be able to represent this time dependence. In standard offline RL datasets, trajectories are stored with timestep information, but each sample's contribution to the loss does not depend on when it occurs within the episode. This allows data augmentation along the time axis, which we describe in the following section.
\begin{algorithm}[htbp]
\caption{AETDICE: Offline AET Optimization via Distribution Correction Estimation}
\label{alg:aetdice}
\begin{algorithmic}[1]
\Require Offline dataset
$\mathcal{D}=\{(s_t,a_t,\mathbf{r}_t,s_{t+1},t)\}$ from the original MOMDP $\mathcal{M}$;
transformation $F:\mathbb{R}^m\!\to\!\mathbb{R}^n$; aggregation $G:\mathbb{R}^n\!\to\!\mathbb{R}$;
dual network $\nu_\psi$, policy network $\tilde{\pi}_\theta$;
divergence temperature $\beta>0$; learning rate $\eta$
\Ensure Augmented-state policy $\tilde{\pi}_\theta$ approximating the AET-optimal policy
 
\Statex \textbf{// Stage 1: Dataset preprocessing (one-time)}
\State Construct augmented dataset $\widetilde{\mathcal{D}}$: for each trajectory in $\mathcal{D}$,
compute accumulated rewards $\mathbf{R}^{\mathrm{acc}}_t = \sum_{k=0}^{t-1}\mathbf{r}(s_k,a_k)$
and form augmented transitions $(\tilde{s}_t,a_t,\tilde{\mathbf{r}}_t,\tilde{s}_{t+1})$ with
$\tilde{s}_t=(s_t,\mathbf{R}^{\mathrm{acc}}_t,t)$
\State Compute the transformed reward for each transition:
\[
    \tilde{\mathbf{r}}_t \;=\; F\!\left(\mathbf{R}^{\mathrm{acc}}_t + \mathbf{r}(s_t,a_t)\right) - F\!\left(\mathbf{R}^{\mathrm{acc}}_t\right) \;\in\; \mathbb{R}^n
\]
 
\Statex \textbf{// Stage 2: Initialization}
\State Initialize $\psi$, $\theta$
\If{$G$ is nonlinear}
    \State Initialize $\boldsymbol{\mu}\in\mathbb{R}^n$
\Else
    \State Fix $\boldsymbol{\mu}=\mathbf{1}$ or preference weights of a choice (no slack constraint needed when $G$ is linear)
\EndIf
 
\Statex \textbf{// Stage 3: Joint dual / policy optimization}
\While{not converged}
    \State Sample minibatch $\mathcal{B}=\{(\tilde{s}_t,a_t,\tilde{\mathbf{r}}_t,\tilde{s}_{t+1})\}$ from $\widetilde{\mathcal{D}}$ and initial states $\tilde{s}_0\sim\tilde{p}_0$
 
    \State Compute empirical slack-augmented advantage at each transition:
    \[
        \hat{e}_{\boldsymbol{\mu},\psi,t}(\tilde{s}_t,a_t) \;=\;
        \boldsymbol{\mu}^{\!\top}\tilde{\mathbf{r}}_t \;+\; \nu_\psi(\tilde{s}_{t+1}) \;-\; \nu_\psi(\tilde{s}_t)
    \]
    \Statex \hspace{1.5em} (single-sample plug-in for
    $\mathbb{E}_{\tilde{s}'\sim\tilde{P}}[\nu_{t+1}(\tilde{s}')]$;
    boundary $\nu_\psi(\tilde{s}_H):=0$ at terminal transitions)
 
    \State Update dual parameters $\psi$ (and $\boldsymbol{\mu}$ if $G$ is nonlinear) by minimizing the empirical AET dual loss:
    \[
        \widehat{\mathcal{L}}_{\mathrm{AET}}(\psi,\boldsymbol{\mu}) \;=\;
        \mathbb{E}_{\tilde{s}_0\sim\tilde{p}_0}\!\big[\nu_\psi(\tilde{s}_0)\big]
        \;+\;
        \mathbb{E}_{(\tilde{s}_t,a_t,\tilde{s}_{t+1})\sim\widetilde{\mathcal{D}}}\!\left[\beta\,\phi^{*}_{+}\!\!\left(\frac{\hat{e}_{\boldsymbol{\mu},\psi,t}(\tilde{s}_t,a_t)}{\beta}\right)\right]
        \;+\;
        G^{*}(-\boldsymbol{\mu})
    \]
    \Statex \hspace{1.5em} via $\psi\leftarrow\psi - \eta \nabla_\psi \widehat{\mathcal{L}}_{\mathrm{AET}}$
    and $\boldsymbol{\mu}\leftarrow\boldsymbol{\mu} - \eta \nabla_{\boldsymbol{\mu}} \widehat{\mathcal{L}}_{\mathrm{AET}}$
 
    \State Compute optimal visitation ratio at sampled transitions:
    \[
        w^{*}(\tilde{s}_t,a_t) \;=\;
        \max\!\left(0,\;(\phi')^{-1}\!\!\left(\frac{\hat{e}_{\boldsymbol{\mu},\psi,t}(\tilde{s}_t,a_t)}{\beta}\right)\right)
    \]
 
    \State Update policy parameters $\theta$ via weighted behavior cloning:
    \[
        \mathcal{L}_{\theta} \;=\;
        -\,\mathbb{E}_{(\tilde{s}_t,a_t)\sim\widetilde{\mathcal{D}}}\!\left[w^{*}(\tilde{s}_t,a_t)\,\log\tilde{\pi}_\theta(a_t\mid\tilde{s}_t)\right]
    \]
    \Statex \hspace{1.5em} via $\theta\leftarrow\theta - \eta \nabla_\theta \mathcal{L}_\theta$
\EndWhile
 
\State \Return $\tilde{\pi}_\theta$
\end{algorithmic}
\end{algorithm}

\section{Time-Index Augmentation in the Augmented MOMDP}
\label{app:time_augmentation}
We describe the data augmentation procedure used to support learning in the augmented MOMDP. The goal is to enrich the dataset with additional time-indexed contexts while preserving the original state-action distribution.
\subsection{Dataset and Augmentation Strategy}
We begin with a dataset $\mathcal{D}_0$ collected by a behavior policy $\pi_b$, consisting of tuples $(s_t, a_t, \mathbf{R}^{\mathrm{acc}}_t, t)$, where $\mathbf{R}^{\mathrm{acc}}_t = \sum_{k=0}^{t-1} \mathbf{r}(s_k,a_k)$ is recorded during data collection. No assumptions are made about the optimality of $\pi_b$. Augmentation is performed solely over the discrete time index $t \in \{0,1,\dots,H\}$; all other components of the augmented state—including $s$ and $\mathbf{R}^{\mathrm{acc}}$—are taken directly from $\mathcal{D}_0$ and are never resampled or synthetically modified.
\subsection{Multi-Head Lagrangian Multiplier}
Augmentation is implemented through the parameterization of the Lagrangian multiplier $\nu(s,\mathbf{R}^{\mathrm{acc}},t)$. The neural network consists of a shared trunk processing $(s,\mathbf{R}^{\mathrm{acc}})$, followed by a multi-head output indexed by time, producing $[\nu(s,\mathbf{R}^{\mathrm{acc}},0), \dots, \nu(s,\mathbf{R}^{\mathrm{acc}},H)]$. For each data point $(s,\mathbf{R}^{\mathrm{acc}})$ from $\mathcal{D}_0$, the full set of time-indexed outputs is available during training. This exposes the dual objective to a richer collection of time-conditioned contexts without altering the underlying state-action distribution, and ensures that the multipliers are well-trained across the entire episode horizon.

\section{Environment Details}
\label{app:env_details}
\begin{figure}[h]
    \centering
    \captionsetup{font=footnotesize}
    \begin{subfigure}[b]{0.32\linewidth}
        \includegraphics[width=\linewidth]{figure/fair_taxi_env_v1.pdf}
        \caption{Fair-Taxi}
        \label{fig:taxi_env}
    \end{subfigure}
    \hfill
    \begin{subfigure}[b]{0.32\linewidth}
        \includegraphics[width=\linewidth]{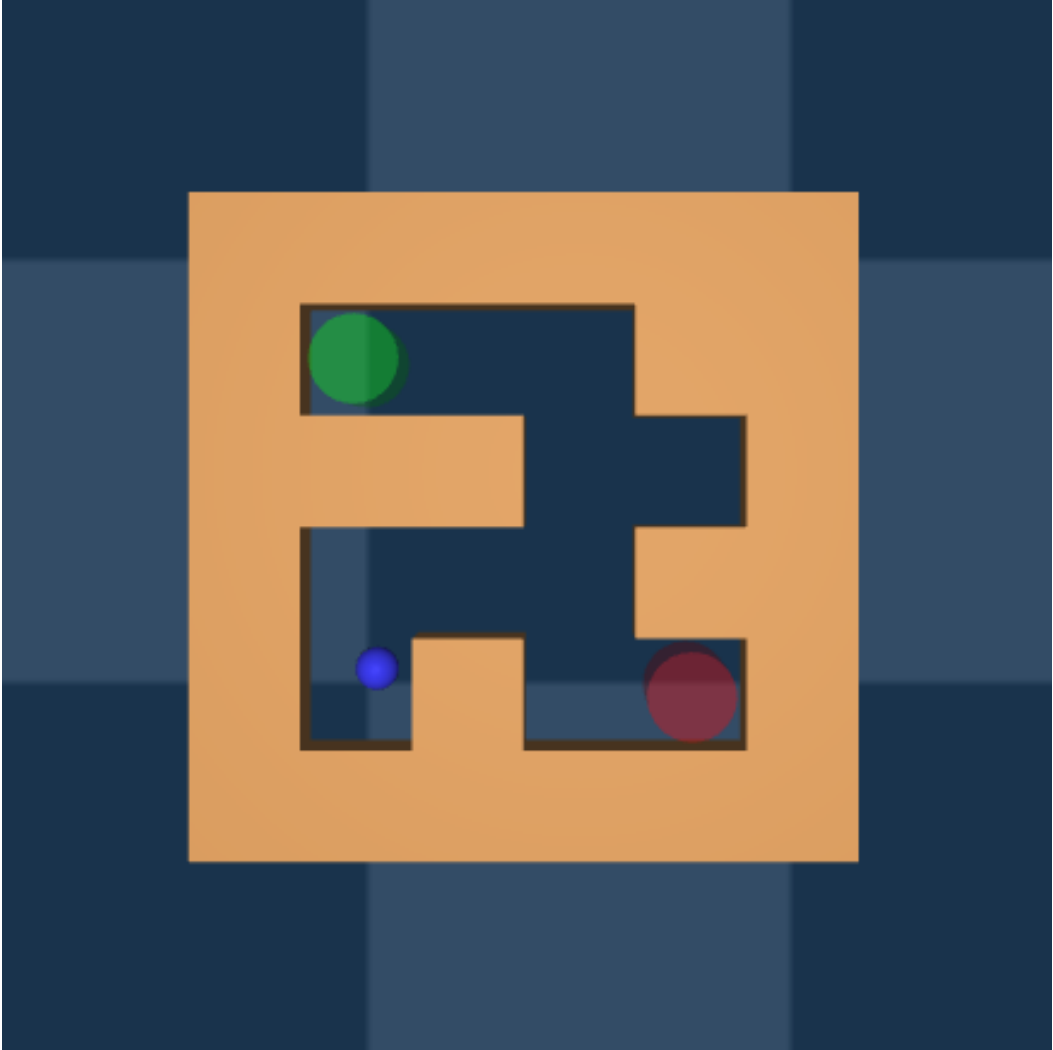}
        \caption{MO-PointMaze-2obj}
        \label{fig:pointmaze_2obj}
    \end{subfigure}
    \hfill
    \begin{subfigure}[b]{0.32\linewidth}
        \includegraphics[width=\linewidth]{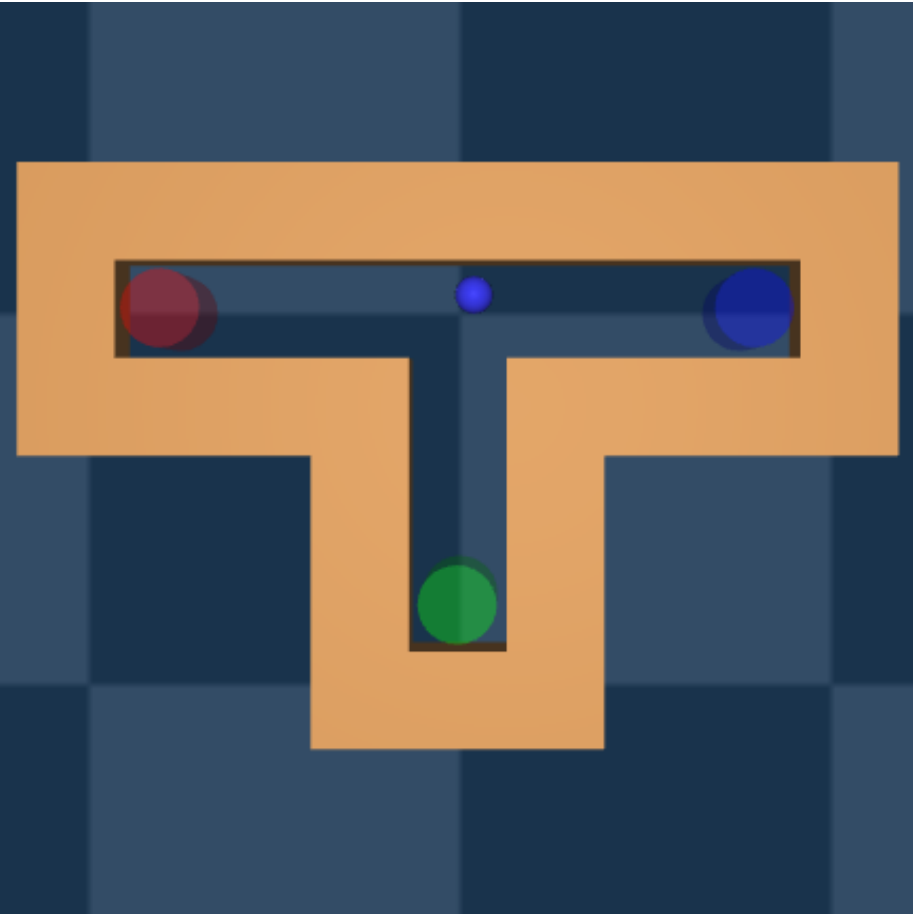}
        \caption{MO-PointMaze-3obj}
        \label{fig:pointmaze_3obj}
    \end{subfigure}
    \caption{
    Visualization of the multi-objective environments. Each environment balances multiple objectives over a finite horizon ($H=50$ for Fair-Taxi and $H=300$  MO-PointMaze).
    }
      \vskip -0.18in
\end{figure}

\subsection{Finite Fair-Taxi}
The Finite Fair-Taxi environment is adapted from the multi-objective taxi domain of Fan et al.~\cite{fan2023welfare} and formulated as a finite-horizon grid-world task. A taxi operates on an $N \times N$ grid and serves passengers associated with different origin--destination pairs, where each passenger group defines one reward dimension. Because only one passenger can be transported at a time and episodes have a fixed horizon, the agent must allocate time across groups, introducing a fairness trade-off between objectives. All experiments use a two-objective setting.

\paragraph{State and Action Spaces.}
The state is discrete and encodes the taxi location $(x,y)$, a binary indicator of whether a passenger is onboard, and the index of the passenger group currently being served (or a special value if none). The action space consists of six discrete actions: move north, south, east, west, pick up, and drop off. Movement actions are clipped at grid boundaries. Pickup is valid only when the taxi is at an origin location and not already carrying a passenger, and drop-off is valid only when the taxi is at the corresponding destination of the onboard passenger.

\paragraph{Reward Function.}
Rewards are vector-valued with dimension $m=2$. At each time step, the reward vector $\mathbf{r}_t \in \mathbb{R}^2$ is defined as follows: a successful delivery for group $i$ yields $r_t^{(i)}=30$ and $r_t^{(j)}=0$ for $j \neq i$; invalid pickup or drop-off actions incur $r_t^{(1)} = r_t^{(2)} = -10$; all other transitions produce zero reward. To introduce asymmetry between objectives, the rewards and penalties for group 1 are scaled by a factor of two, resulting in $r_t^{(1)} \in \{60, -20, 0\}$ while group 2 retains $\{30, -10, 0\}$. Rewards are therefore sparse and event-driven.

\paragraph{Episode Termination and Layout.}
Episodes terminate after a fixed horizon of 50 time steps. The two origin--destination pairs are placed in distant regions of the grid, with each origin uniquely paired to a destination. Serving one group typically requires traveling in a direction that does not overlap with the other, so dedicating time to one group reduces the opportunity to serve the other. This creates a temporal trade-off between objectives within each episode.

\paragraph{Offline Dataset Construction}
For the Fair-Taxi domain, offline datasets are generated using a mixture of single-objective policies following the strategy. Each policy is a tabular Q-function optimized for a different linear scalarization of the reward vector. During data collection, policies are alternated within each episode according to fixed time intervals, producing trajectories that contain segments optimized for different objectives. This switching scheme encourages diverse return profiles across episodes and prevents the dataset from being dominated by trajectories specialized to a single objective. Episodes start from a fixed initial taxi location to ensure comparability across policies. Each episode runs for at most 50 steps. The dataset stores state, action, next state, reward vector, timestep index, terminal flag, and accumulated return vector at each step.

\subsection{MO-PointMaze}

The MO-PointMaze environment is a continuous-state multi-objective navigation task adapted from the D4RL maze domains~\cite{fu2020d4rl}. A point-mass agent moves in a 2D maze containing multiple spatial goal regions, where each goal corresponds to one reward dimension. Because the goals are spatially separated by walls and corridors, the agent must allocate time across regions, inducing a trade-off between objectives within each episode.

\paragraph{State and Action Spaces.}
The observation space is inherited from the underlying D4RL PointMaze environment and includes the agent’s 2D position and velocity. Experiments use a discretized action space consisting of movements along the four cardinal directions with several fixed magnitudes. This discretization stabilizes training and ensures consistent coverage during offline data collection. The episode horizon is set to 300 steps.

\paragraph{Maze Layouts.}
Two layouts are used. In the two-objective setting~\ref{fig:pointmaze_2obj}, the agent starts bottom left region of a compact maze with two goal regions located in different corridors. In the three-objective setting~\ref{fig:pointmaze_3obj}, the maze contains three goals positioned in distinct regions. Reaching one goal generally requires moving away from the others, making simultaneous proximity infeasible.

\paragraph{Reward Function.}
Rewards are vector-valued with dimension equal to the number of goals $m$. At each time step, the reward for objective $i$ depends on the Euclidean distance between the agent’s position $\mathbf{s}_t$ and the $i$-th goal location $\mathbf{g}_i$:
\[
r_t^{(i)} = \exp\!\left(- \frac{\lVert \mathbf{s}_t - \mathbf{g}_i \rVert}{d_0} \right),
\]
where $d_0$ is a scaling constant proportional to the maze cell size. Rewards decay rapidly with distance, so meaningful reward in one dimension requires the agent to remain near the corresponding goal region for a sustained period.

In the two-objective setting, the first reward dimension is scaled by a factor of two to introduce asymmetry between objectives. In the three-objective setting, all reward dimensions are equally scaled.

\paragraph{Multi-Objective Trade-off.}
Due to the rapid spatial decay of rewards and the maze topology, the agent cannot remain close to multiple goals at the same time. Policies must therefore schedule visits to different regions within an episode, creating a temporal trade-off between objectives. This makes the environment well suited for analyzing differences between trajectory-level and expectation-level scalarization.

\begin{figure} 
    \centering 
    \includegraphics[width=1.0\linewidth]{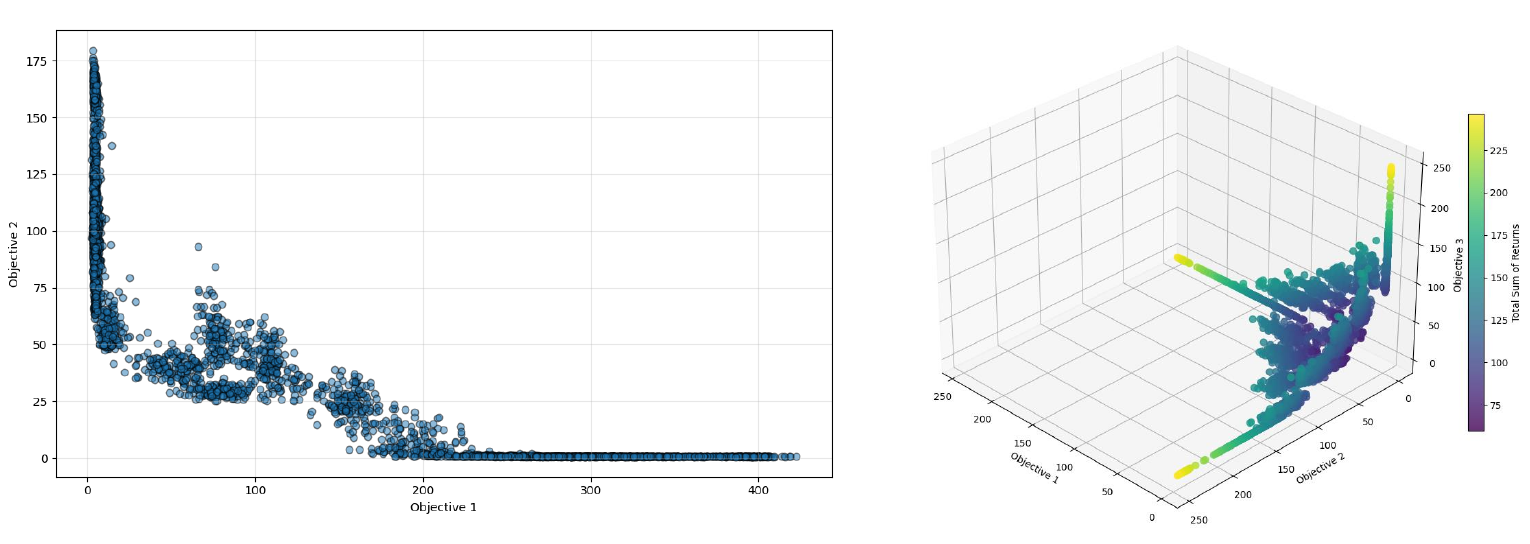} 
    \caption{Visualization of the offline dataset return distributions for the MO-PointMaze environment. \textbf{(Left)} The cumulative returns in the 2-objective setting, illustrating the trade-off between the two goals. \textbf{(Right)} The spatial distribution of returns in the 3-objective setting.} 
    \label{fig:dataset}
\end{figure}

\paragraph{Offline Dataset Collection.}
Offline datasets are generated using heuristic navigation policies that move toward specific goal regions following shortest-path routes. By varying goal targets and switching patterns over time, the dataset covers diverse behaviors, including specialization on a single objective and sequential visitation of multiple goals. This diversity ensures that the dataset contains trajectories with a wide range of return trade-offs, which is essential for evaluating multi-objective offline RL methods. The resulting return distributions are visualized in Figure~\ref{fig:dataset}.

\subsection{MO-MuJoCo (D4MORL)}

We use the multi-objective MuJoCo benchmark from D4MORL~\citep{zhu2023scaling},
which extends the standard MuJoCo locomotion suite by replacing scalar rewards
with vector-valued multi-objective rewards while leaving the underlying physics,
observation space, and action space unchanged. Five environments are used:
MO-Ant, MO-Walker2d, MO-HalfCheetah. For all
environments we use the \emph{expert-uniform} and \emph{amateur-uniform} dataset
splits provided by~\cite{zhu2023scaling}, in which the behaviour-policy
preferences are sampled uniformly from the 2-simplex during data collection;
each split contains 50,000 transitions per env.

\paragraph{State and Action Spaces.}
States and continuous actions are inherited from the underlying MuJoCo
environments. State and action dimensions are Ant 27/8, HalfCheetah 17/6, Walker2d 17/6, Hopper 11/3, Swimmer 8/2 --- verify.
We follow the standard MuJoCo convention of clipping actions to $[-1, 1]$
before computing the reward.

\paragraph{Reward Function.}
At each time step the reward is a vector $\mathbf{r}_t \in \mathbb{R}^{2}$ whose
two components capture distinct physical aspects of the task: forward and lateral
velocity in MO-Ant, forward speed and energy saving in MO-HalfCheetah and
MO-Walker2d, forward speed and energy efficiency in MO-Swimmer, and running
and jumping in MO-Hopper. Both components include a shared survival bonus and a
shared action-norm penalty, so the two objectives are positively oriented
(higher is better) and partially correlated. We refer to~\cite{zhu2023scaling}
for exact formulae. To study the effect of asymmetry between objectives, the
per-objective rewards are min--max normalized at dataset load time:
\[
\tilde{r}_t^{(i)} \;=\; c_i \cdot \frac{r_t^{(i)} - r_{\min}^{(i)}}{r_{\max}^{(i)} - r_{\min}^{(i)}},
\]
where $r_{\min}^{(i)}, r_{\max}^{(i)}$ are computed over the offline dataset and
$c_i \in \mathbb{R}_+$ is a per-objective scaling factor. We use the symmetric
default $\mathbf{c} = (1,1)$ for all environments and additionally
$\mathbf{c} = (2,1)$ in MO-Ant for an objective-1-favouring variant.

\paragraph{Episode Termination and Horizon.}
All environments use a fixed horizon of $H=500$ steps. Standard MuJoCo
early-termination conditions apply (height/angle bounds in MO-Walker2d and
MO-Hopper, instability bounds in MO-Ant); transitions following early
termination contribute zero reward and are masked out by the buffer.

\paragraph{Cobb-Douglas Reward Reshaping.}
Two of our experiments (G-linear with $F$-Cobb-Douglas, and G-log with combined
$F$-Cobb-Douglas and $F$-threshold) require the reward vector to be split into a
clean (return, cost) pair. The relevant D4MORL environments admit such a split
because their second objective takes the saving form
$\text{const} - \|\mathbf{a}\|^{2} + \text{alive}$. For MO-HalfCheetah, we
recover the action-norm cost as $r_t^{(1)} \leftarrow 5 - r_t^{(1)}$, yielding a
$(\text{forward speed},\; \|\mathbf{a}\|^{2})$ layout. For MO-Walker2d we use a
3-dimensional layout $(\text{speed},\, \text{cost},\, \text{safety})$, where
speed and cost match the MO-HalfCheetah reshape and the safety channel is a
quadratic balance reward
$r_t^{\text{safety}} = (1 - 2 h_{\text{norm}}^{2}) + (1 - 2 a_{\text{norm}}^{2})$
computed from the height and torso-angle observations normalized to the env's
safety bounds (so $r_t^{\text{safety}} \in [-2, +2]$ in-bounds, peaking at the
upright centre). The Cobb-Douglas utility consumes (speed, cost), and the
safety channel feeds a separate piecewise-log threshold; the two combine
through the outer aggregation~$G$.

\section{Additional Experiment Results}
\label{app:additional_exp}

\subsection{Fair-Taxi}


We report quantitative performance on the Finite Fair-Taxi environment under ESR, BSR ($\alpha=0.5$), and SER objectives.
All values are log-scaled utilities (mean $\pm$ standard deviation over three seeds), so numerical differences reflect multiplicative gaps in the underlying objective values.
Each policy is trained using the same AET framework with a different nonlinear objective and evaluated under all three criteria; entry $(i,j)$ in Table~\ref{sample-table-indented} reports the performance of a policy trained for objective $i$ when evaluated using objective $j$.

Across all objectives, policies achieve their highest performance when evaluated under the same criterion used during training, indicating that ESR, BSR, and SER correspond to distinct optimization targets in practice.
Policies trained under BSR exhibit intermediate behavior: while optimized for BSR, they retain comparatively higher performance under ESR and SER evaluations than policies trained under the extreme objectives.
Overall, these results demonstrate that nonlinear objectives are not interchangeable, and that objective mismatch can lead to substantial degradation even when numerical differences appear small due to log scaling.

\begin{table}[h]
  \caption{Performance of offline Fair MORL in Fair-Taxi environment.
Rows correspond to the nonlinear objective used for training, and columns correspond to the evaluation objective.
All values are log-scaled utilities (mean $\pm$ standard deviation over 3 random seeds).}
  \label{sample-table-indented}
  \begin{center}
    \begin{small}
      \begin{sc}
        \begin{tabular}{l|ccc}
          \toprule
          Method & ESR & BSR & SER \\
          \midrule
            AET-ESR& $\mathbf{1.197} {\scriptstyle \pm0.002}$ & $-0.797 {\scriptstyle \pm 0.002}$ & $1.130 {\scriptstyle \pm 0.000}$ \\
            AET-BSR& $0.193 {\scriptstyle \pm 0.011}$ & $\mathbf{-0.788} {\scriptstyle \pm 0.004}$ & $1.748 {\scriptstyle \pm 0.009}$ \\
            AET-SER& $0.110 {\scriptstyle \pm 0.034}$ & $-1.045 {\scriptstyle \pm 0.035}$ & $\mathbf{1.804} {\scriptstyle \pm 0.002}$ \\
          \bottomrule
        \end{tabular}
      \end{sc}
    \end{small}
  \end{center}
  \vskip -0.1in
\end{table}

\subsection{MO-PointMaze}
\begin{figure}[!htbp]
  \vskip -0.0in
  \begin{center}
    \centerline{\includegraphics[width=0.7\columnwidth]{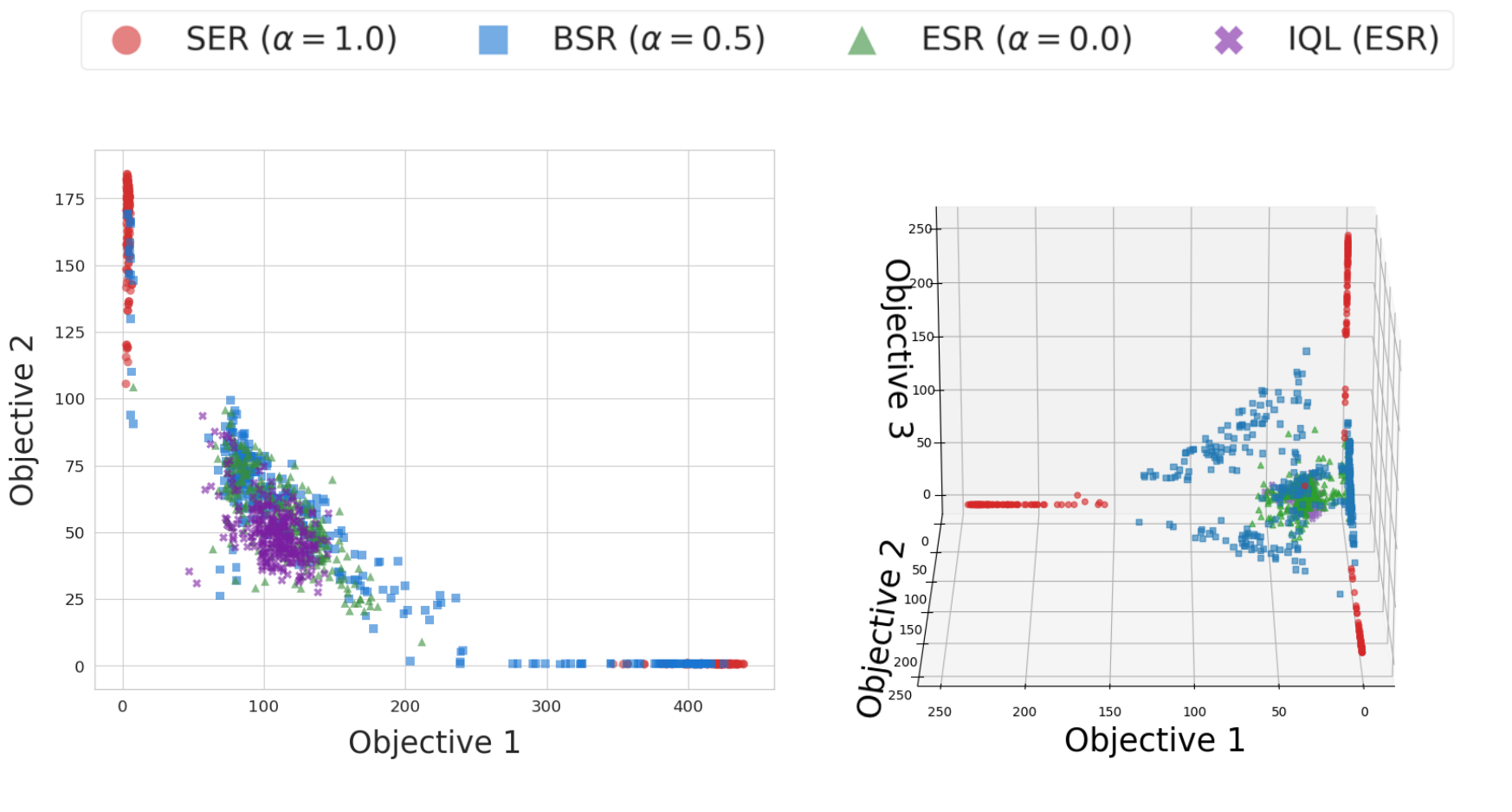}}
    \caption{Performance results in the MO-PointMaze-2obj (left) and MO-PointMaze-3obj (right) environment.}
    \label{pointmaze_evaluaton}
  \end{center}
  \vskip -0.3in
\end{figure}

We next examine how these behavioral differences translate into return trade-offs in a continuous-state setting. Figure~\ref{pointmaze_evaluaton} shows the achieved return vectors. In the two-objective case, ESR produces balanced trajectories along the diagonal trade-off, whereas SER concentrates near extreme regions that prioritize a single objective per episode. BSR ($\alpha=0.5$) spans intermediate trade-offs. The same pattern is more pronounced in the three-objective setting, where SER lies near the corners, ESR occupies the balanced interior, and BSR fills the space between. ESR-IQL aligns with ESR-style trade-offs, while AETDICE covers the full spectrum.

Table~\ref{tab:full-comparison-stacked} reports performance under multiple evaluation criteria (ESR, BSR, SER, and LSR), reflecting different preference structures. The AETDICE variants achieve strong performance on their respective target objectives, demonstrating that the proposed framework can effectively optimize a range of nonlinear criteria within a unified approach. In the three-objective setting, ESR-IQL attains the highest ESR score, consistent with its specialization. Linear scalarization achieves high linear scalarized return (LSR), as expected, but yields lower performance under nonlinear utilities, often producing policies that prioritize a single objective rather than balancing across objectives.


\begin{table}[t] 
\caption{Comparison of ESR, BSR, SER, and LSR (linear scalarization return with coefficients of 1) scores in MO-PointMaze. For each method, performance is reported with respect to the objective it directly optimizes (mean $\pm$ standard deviation over 3 random seeds).}
\label{tab:full-comparison-stacked}
\begin{center}
\begin{small}
\begin{sc}
    \begin{tabular}{l cccc} 
        \toprule
        Method & ESR & $\text{BSR}_{0.5}$ & SER & LSR \\
        \midrule
        
        \multicolumn{5}{l}{\textbf{MO-PointMaze-2obj}} \\
        \midrule
        \textit{AETDICE} \\
        \hspace{2mm}ESR & $\mathbf{6.67} {\scriptstyle \pm 0.08}$ & $11.90 {\scriptstyle \pm 0.17}$ & $8.73 {\scriptstyle \pm 0.08}$ & $164.86 {\scriptstyle \pm 6.68}$ \\
        \hspace{2mm}$\text{BSR}_{0.5}$ & $5.91 {\scriptstyle \pm 0.32}$ & $\mathbf{11.93} {\scriptstyle \pm 0.11}$ & $9.01 {\scriptstyle \pm 0.04}$ & $211.79 {\scriptstyle \pm 12.29}$ \\
        \hspace{2mm}SER & $4.06 {\scriptstyle \pm 0.05}$ & $11.82 {\scriptstyle \pm 0.05}$ & $\mathbf{9.74} {\scriptstyle \pm 0.02}$ & $284.05 {\scriptstyle \pm 1.42}$ \\
        \addlinespace[2pt]
        \textit{Linear Scalarization} \\
        \hspace{2mm}$\mathbf{w}=[0.5,0.5]$ & $3.64 {\scriptstyle \pm 0.01}$ & $8.77 {\scriptstyle \pm 0.02}$ & $3.81 {\scriptstyle \pm 0.02}$ & $\mathbf{417.01} {\scriptstyle \pm 3.13}$ \\
        \hspace{2mm}$\mathbf{w}=[1.0,0.0]$ & $3.63 {\scriptstyle \pm 0.02}$ & $8.71 {\scriptstyle \pm 0.04}$ & $3.82 {\scriptstyle \pm 0.03}$ & $409.62 {\scriptstyle \pm 5.70}$ \\
        \hspace{2mm}$\mathbf{w}=[0.75,0.25]$ & $3.67 {\scriptstyle \pm 0.02}$ & $8.91 {\scriptstyle \pm 0.19}$ & $4.53 {\scriptstyle \pm 1.19}$ & $415.88 {\scriptstyle \pm 6.09}$ \\
        \addlinespace[2pt]
        \textit{Other} \\
        \hspace{2mm}ESR-IQL & $6.56 {\scriptstyle \pm 0.01}$ & $11.59 {\scriptstyle \pm 0.03}$ & $8.56 {\scriptstyle \pm 0.02}$ & $153.78 {\scriptstyle \pm 1.49}$ \\
        
        \midrule \addlinespace[4pt]

        \multicolumn{5}{l}{\textbf{MO-PointMaze-3obj}} \\
        \midrule
        \textit{AETDICE} \\
        \hspace{2mm}ESR & $6.79 {\scriptstyle \pm 0.15}$ & $11.76 {\scriptstyle \pm 0.10}$ & $10.15 {\scriptstyle \pm 0.02}$ & $86.05 {\scriptstyle \pm 0.46}$ \\
        \hspace{2mm}$\text{BSR}_{0.5}$ & $4.72 {\scriptstyle \pm 0.24}$ & $\mathbf{12.04} {\scriptstyle \pm 0.07}$ & $11.14 {\scriptstyle \pm 0.05}$ & $120.34 {\scriptstyle \pm 1.77}$ \\
        \hspace{2mm}SER & $-0.19 {\scriptstyle \pm 0.03}$ & $11.43 {\scriptstyle \pm 0.03}$ & $\mathbf{12.81} {\scriptstyle \pm 0.02}$ & $\mathbf{213.78} {\scriptstyle \pm 1.38}$ \\
        \addlinespace[2pt]
        \textit{Linear Scalarization} \\
        \hspace{2mm}$\mathbf{w}=[0.5,0.5]$ & $-0.26 {\scriptstyle \pm 0.07}$ & $10.61 {\scriptstyle \pm 0.06}$ & $12.27 {\scriptstyle \pm 0.01}$ & $176.92 {\scriptstyle \pm 1.63}$ \\
        \hspace{2mm}$\mathbf{w}=[1.0,0.0]$ & $-0.34 {\scriptstyle \pm 0.03}$ & $2.49 {\scriptstyle \pm 0.06}$ & $2.17 {\scriptstyle \pm 0.03}$ & $203.20 {\scriptstyle \pm 4.01}$ \\
        \hspace{2mm}$\mathbf{w}=[0.75,0.25]$ & $-0.38 {\scriptstyle \pm 0.01}$ & $2.41 {\scriptstyle \pm 0.04}$ & $-2.20 {\scriptstyle \pm 0.03}$ & $198.15 {\scriptstyle \pm 3.22}$ \\
        \addlinespace[2pt]
        \textit{Other} \\
        \hspace{2mm}ESR-IQL & $\mathbf{7.20} {\scriptstyle \pm 0.03}$ & $12.03 {\scriptstyle \pm 0.04}$ & $10.17 {\scriptstyle \pm 0.02}$ & $87.16 {\scriptstyle \pm 0.94}$ \\
        \bottomrule
    \end{tabular}
\end{sc}
\end{small}
\end{center}
\end{table}

\subsection{D4MORL}
\paragraph{Environments and reward modifications.}
We run three experiment families on D4MORL. The first is experiment on the performance of various AETDICE objectives on
MO-Ant and MO-Walker2d with the original D4MORL rewards and datasets. 
The second is the Cobb--Douglas (CD) experiment on MO-Walker2d
and MO-HalfCheetah, where we reshape the reward into a (return, cost) layout because the CD utility requires a monotone-increasing reward in dimension~0 and a monotone-increasing cost in dimension~1. 
The third is the safety utility aggregated CD experiment on MO-Walker2d, which adds a safety reward to prevent a ``drop-and-die'' exploit.

\textbf{CD reshape (MO-Walker2d, MO-HalfCheetah).} Both environments expose
a forward-speed reward $r^{v} = v_{x} + r^{\text{alive}}$ and an
energy-saving reward $r^{e} = 4 - \|a\|^{2} + r^{\text{alive}}$. We keep the
forward speed unchanged and recover the action-energy cost from $r^{e}$:
\[
  R_{0} \;=\; r^{v}, \qquad R_{1} \;=\; 5 - r^{e} \;=\; \|a\|^{2}.
\]

\textbf{Safety utility aggregated CD (MO-Walker2d).} Walker2d ends an episode when the torso height $h$ leaves $[0.8, 2.0]$ or the torso angle $\theta$ leaves
$[-1, 1]$. A CD agent at low~$\rho$ can game this by ending the episode early to shrink the cost denominator. To prevent this we add a third reward that scores the posture or location of the upper body of walker2d agent inside the safe zone. Normalizing $h$ and
$\theta$ so the safe zone maps to $[-1, 1]$ via $\bar h = (h - 1.4)/0.6$
and $\bar\theta = \theta$, we define
\[
  R_{2} \;=\; \bigl(-2\,\bar h^{2} + 1\bigr) \;+\; \bigl(-2\,\bar\theta^{2} + 1\bigr).
\]
Per step, $R_{2} \in [-2, 2]$ inside the safe zone, peaking at $+2$ at the
upright center and dropping to $-2$ at the env bounds. $R_{2}$ feeds the
threshold branch of the $F$-utility while $R_{0}$ and $R_{1}$ feed the
Cobb--Douglas branch, so the safety floor is decoupled from the
speed/cost trade-off.

\begin{table}[!htbp]
  \caption{Results of MO-Ant Expert with Concave G and Convex F. (mean ± standard deviation over 5 random seeds)}
  \label{tab:results_ant_expert}
  \begin{center}
    \begin{small}
      \begin{sc}
        \begin{tabular}{l|c|c|c|c}
          \toprule
          Method & $w=[1,0]$ & $w=[0.5,0.5]$ & $w=[0,1]$ & $G$, $F$ \\
          \midrule
            Lin-$[1.0, 0.0]$& $1089.56 \pm 27.60$ & $729.61 \pm 10.94$ & $369.66 \pm 11.93$ & $12.91 \pm 0.03$\\
            Lin-$[0.5, 0.5]$& $584.69 \pm 18.29$ & $689.23 \pm 12.74$ & $793.77 \pm 23.10$ & $13.05 \pm 0.04$\\
            Lin-$[0.0, 1.0]$& $88.83 \pm 1.53$ & $1221.97 \pm 26.71$ & $2355.10 \pm 53.65$ & $12.25 \pm 0.03$\\
          FairDICE& $617.17 \pm 19.68$ & $682.65 \pm 20.06$ & $748.13 \pm 30.64$ & $13.04 \pm 0.06$\\
          \midrule
          ESR-IQL\\
            \hspace{5mm}$[1.0, 0.0]$& $\mathbf{1296.49 \pm 30.66}$ & $811.82 \pm 14.71$ & $327.15 \pm 6.20$ & $12.96 \pm 0.03$\\
            \hspace{5mm}$[0.5, 0.5]$& $85.54 \pm 1.61$ & $\mathbf{1243.33 \pm 17.07}$ & $\mathbf{2401.12 \pm 34.37}$ & $12.23 \pm 0.02$\\
            \hspace{5mm}$[0.0, 1.0]$& $87.32 \pm 3.82$ & $1240.07 \pm 19.36$ & $2392.82 \pm 40.01$ & $12.25 \pm 0.04$\\
          AET-DICE& $816.78 \pm 61.16$ & $949.73 \pm 22.02$ & $1082.68 \pm 85.08$ & $\mathbf{13.69 \pm 0.05}$\\
          \bottomrule
        \end{tabular}
      \end{sc}
    \end{small}
  \end{center}
\end{table}

\begin{table}[!htbp]
  \caption{Results of MO-Ant Amateur with Concave G and Convex F. (mean ± standard deviation over 5 random seeds)}
  \label{tab:results_ant_amateur}
  \begin{center}
    \begin{small}
      \begin{sc}
        \begin{tabular}{l|c|c|c|c}
          \toprule
          Method & $w=[1,0]$ & $w=[0.5,0.5]$ & $w=[0,1]$ & $G$, $F$ \\
          \midrule
            Lin-$[1.0, 0.0]$& $869.42 \pm 27.99$ & $636.55 \pm 10.38$ & $403.67 \pm 17.70$ & $12.77 \pm 0.03$\\
            Lin-$[0.5, 0.5]$& $540.80 \pm 29.59$ & $623.84 \pm 13.05$ & $706.89 \pm 29.54$ & $12.85 \pm 0.04$\\
            Lin-$[0.0, 1.0]$& $138.07 \pm 6.21$ & $875.34 \pm 22.25$ & $1612.61 \pm 50.02$ & $12.31 \pm 0.02$\\
          FairDICE& $509.52 \pm 12.81$ & $638.80 \pm 26.01$ & $768.09 \pm 54.61$ & $12.87 \pm 0.07$\\
          \midrule
            \hspace{5mm}$[1.0, 0.0]$& $\mathbf{1570.58 \pm 17.03}$ & $876.82 \pm 9.37$ & $183.06 \pm 3.94$ & $12.57 \pm 0.03$\\
            \hspace{5mm}$[0.5, 0.5]$& $98.57 \pm 14.41$ & $976.31 \pm 13.21$ & $1854.05 \pm 37.46$ & $12.11 \pm 0.12$\\
            \hspace{5mm}$[0.0, 1.0]$& $88.11 \pm 2.58$ & $\mathbf{977.25 \pm 13.05}$ & $\mathbf{1866.40 \pm 27.70}$ & $12.01 \pm 0.02$\\
          AET-DICE& $610.17 \pm 46.86$ & $654.87 \pm 7.66$ & $699.56 \pm 31.71$ & $\mathbf{12.96 \pm 0.03}$\\
          \bottomrule
        \end{tabular}
      \end{sc}
    \end{small}
  \end{center}
\end{table}

\begin{table}[!htbp]
  \caption{Results of MO-Walker2d Expert with Concave G and Convex F. (mean ± standard deviation over 5 random seeds)}
  \label{tab:results_walker_expert}
  \begin{center}
    \begin{small}
      \begin{sc}
        \begin{tabular}{l|c|c|c|c}
          \toprule
          Method & $w=[1,0]$ & $w=[0.5,0.5]$ & $w=[0,1]$ & $G$, $F$ \\
          \midrule
            Lin-$[1.0, 0.0]$& $\mathbf{718.58 \pm 23.50}$ & $675.69 \pm 40.14$ & $632.80 \pm 93.83$ & $13.02 \pm 0.13$\\
            Lin-$[0.5, 0.5]$& $279.96 \pm 41.59$ & $1856.50 \pm 202.64$ & $3433.03 \pm 443.57$ & $13.76 \pm 0.06$\\
            Lin-$[0.0, 1.0]$& $10.28 \pm 1.19$ & $7292.20 \pm 402.27$ & $14574.12 \pm 805.61$ & $11.91 \pm 0.07$\\
          FairDICE& $307.14 \pm 15.98$ & $1733.19 \pm 97.94$ & $3159.25 \pm 207.63$ & $13.78 \pm 0.04$\\
          \midrule
            \hspace{5mm}$[1.0, 0.0]$& $689.95 \pm 85.09$ & $772.39 \pm 123.46$ & $854.84 \pm 328.36$ & $13.21 \pm 0.25$\\
            \hspace{5mm}$[0.5, 0.5]$& $10.53 \pm 1.45$ & $\mathbf{7366.92 \pm 326.81}$ & $\mathbf{14723.31 \pm 654.99}$ & $11.94 \pm 0.10$\\
            \hspace{5mm}$[0.0, 1.0]$& $12.35 \pm 0.83$ & $6765.08 \pm 427.43$ & $13517.81 \pm 854.75$ & $12.02 \pm 0.10$\\
          AET-DICE& $246.26 \pm 70.02$ & $2566.75 \pm 850.14$ & $4887.23 \pm 1637.10$ & $\mathbf{13.87 \pm 0.78}$\\
          \bottomrule
        \end{tabular}
      \end{sc}
    \end{small}
  \end{center}
\end{table}
\begin{table}[!htbp]
  \caption{Results of MO-Walker2d Amateur with Concave G and Convex F. (mean ± standard deviation over 5 random seeds)}
  \label{tab:results_walker_amateur}
  \begin{center}
    \begin{small}
      \begin{sc}
        \begin{tabular}{l|c|c|c|c}
          \toprule
          Method & $w=[1,0]$ & $w=[0.5,0.5]$ & $w=[0,1]$ & $G$, $F$ \\
          \midrule
            Lin-$[1.0, 0.0]$& $\mathbf{540.87 \pm 11.01}$ & $491.55 \pm 24.05$ & $442.23 \pm 44.79$ & $12.38 \pm 0.11$\\
            Lin-$[0.5, 0.5]$& $227.71 \pm 13.15$ & $2780.73 \pm 199.93$ & $5333.75 \pm 402.39$ & $14.01 \pm 0.09$\\
            Lin-$[0.0, 1.0]$& $8.37 \pm 0.86$ & $\mathbf{8119.82 \pm 251.87}$ & $\mathbf{16231.27 \pm 504.44}$ & $11.81 \pm 0.08$\\
          FairDICE& $261.07 \pm 18.87$ & $1143.11 \pm 77.84$ & $2025.16 \pm 160.95$ & $13.17 \pm 0.09$\\
          \midrule
            \hspace{5mm}$[1.0, 0.0]$& $176.32 \pm 25.68$ & $149.61 \pm 26.89$ & $122.91 \pm 31.93$ & $9.94 \pm 0.41$\\
            \hspace{5mm}$[0.5, 0.5]$& $42.31 \pm 11.21$ & $6710.95 \pm 614.79$ & $13379.58 \pm 1237.66$ & $13.20 \pm 0.22$\\
            \hspace{5mm}$[0.0, 1.0]$& $39.94 \pm 6.93$ & $6607.13 \pm 427.73$ & $13174.33 \pm 861.51$ & $13.16 \pm 0.12$\\
          AET-DICE& $277.03 \pm 42.64$ & $3376.26 \pm 532.50$ & $6475.49 \pm 1102.24$ & $\mathbf{14.37 \pm 0.08}$\\
          \bottomrule
        \end{tabular}
      \end{sc}
    \end{small}
  \end{center}
\end{table}

\begin{table}[h]
  \caption{Results of Walker2d Cobb-Douglas objectives.}
  \label{tab:results_walker_cd}
  \begin{center}
    \begin{small}
      \begin{sc}
        \begin{tabular}{l|c|c|c|c}
          \toprule
          Method & CD(0.1) & CD(0.5) & safe-CD(0.1) & safe-CD(0.5) \\
          \midrule
          BC & $0.051 \pm 0.043$ & $1.766 \pm 0.453$ & $15.440$ & $18.680$\\
          ESR-IQL& $0.128 \pm 0.120$ & $4.136 \pm 0.728$ & $15.440$ & $19.571$\\
          ESR-DICE& $0.107 \pm 0.060$ & $4.432 \pm 1.270$ & $15.446$ & $19.221$\\
          \midrule
          safe-CD AET& $\mathbf{0.330 \pm 0.043}$ & $\mathbf{4.456 \pm 0.826}$ & $\mathbf{16.272}$ & $\mathbf{19.572}$\\
          \bottomrule
        \end{tabular}
      \end{sc}
    \end{small}
  \end{center}
\end{table}

\begin{table}[h]
  \caption{Results of HalfCheetah Cobb-Douglas objectives.}
  \label{tab:results_halfcheetah}
  \begin{center}
    \begin{small}
      \begin{sc}
        \begin{tabular}{l|c|c}
          \toprule
          Method & CD(0.1) & CD(0.5) \\
          \midrule
          BC & $0.139 \pm 0.092$ & $3.686 \pm 0.838$\\
          ESR-IQL& $1.516 \pm 0.561$ & $22.010 \pm 6.660$\\
          ESR-DICE& $\mathbf{12.842 \pm 2.570}$ & $\mathbf{45.768 \pm 6.580}$\\
          \bottomrule
        \end{tabular}
      \end{sc}
    \end{small}
  \end{center}
\end{table}

\section{Ablation Study: Impact of Regularization Strength}
\label{app:ablation_beta}
\begin{figure}[!htbp]
    \centering 
    \includegraphics[width=1.0\linewidth]{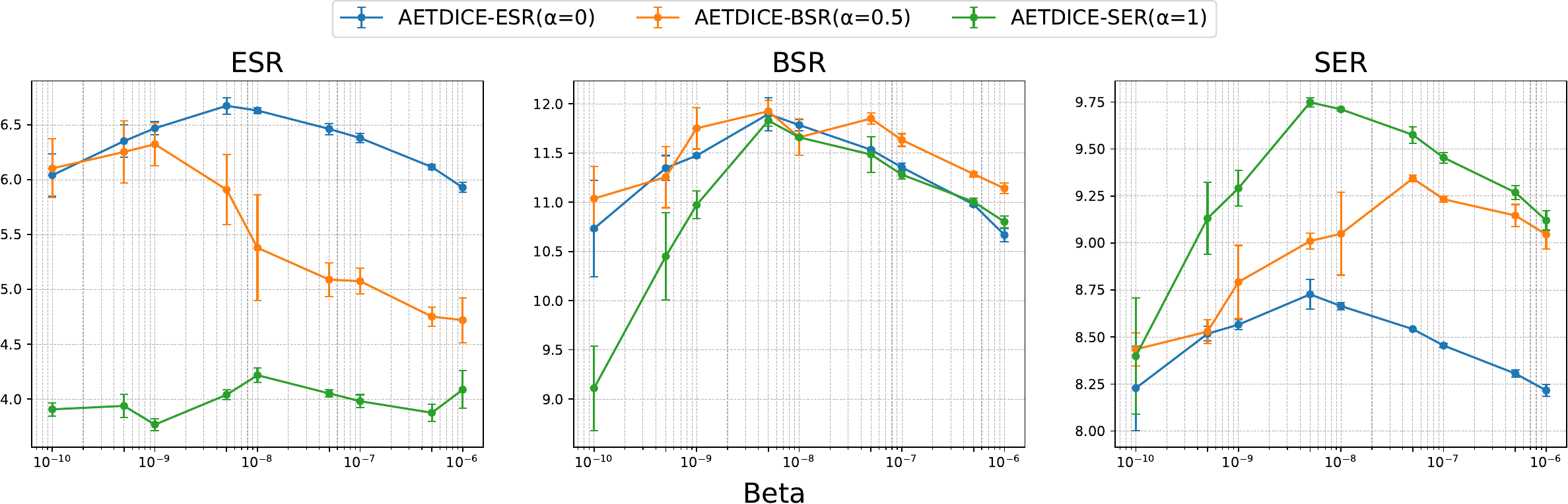} 
    \caption{Performance of AETDICE with varying $\beta$ values on MO-PointMaze-2obj dataset.} 
    \label{fig:beta_ablation} \vskip -0.2in 
\end{figure}
We study how the regularization coefficient $\beta$, which controls the strength of the $\phi$-divergence penalty in AETDICE, affects performance under different AET objectives. Figure~\ref{fig:beta_ablation} reports evaluation scores under ESR, BSR, and SER criteria as $\beta$ varies on MO-PointMaze-2obj.
The effective range of $\beta$ in our experiments is considerably smaller than values commonly reported for DICE-style methods in single-objective settings. This is expected: offline MORL datasets are collected from policies trained under diverse preference weights, so trajectories optimal for one trade-off can be highly suboptimal for another AET objective. Smaller regularization is needed to permit the larger distribution shift required for finding the target objective's optimum within such heterogeneous data. This makes achieving strong offline performance more demanding, as the optimizer must deviate further from the data distribution. Nevertheless, with sufficiently small $\beta$, AETDICE achieves strong performance across all objectives, and performance varies smoothly as $\beta$ changes, indicating stable optimization. These results demonstrate that AETDICE enables effective optimization of AET objectives—including those previously intractable—given appropriate regularization.

\section{Hyperparameter}
\label{app:implementation}

Table~\ref{tab:aetdice_hparams} summarizes the default hyperparameters used for AETDICE. The policy $\pi_\theta$, critic $\nu_\psi$, and preference parameters $\mu$ are all optimized using Adam with a learning rate of $3\times10^{-4}$ and a batch size of 256. Both the policy and critic are implemented as three-layer multilayer perceptrons with hidden dimension 768, except in MO-PointMaze-3obj where we increase the width to 1024 to accommodate the higher-dimensional return structure. Training is performed for $10^6$ gradient steps.

We use the Pearson $\chi^2$ divergence for occupancy regularization. The divergence coefficient $\beta$ is tuned over a logarithmic range $\beta \in \{10^{-6}, 5 \times 10^{-7}, 10^{-7}, 5 \times 10^{-8}, 10^{-8}, 5 \times 10^{-9}, 10^{-9}, 5 \times 10^{-10}, 10^{-10}\}$. This range is smaller than what is typically used in single-objective DICE settings, as nonlinear multi-objective scalarization and trajectory-level return conditioning change the effective scale of the dual residuals. Using a log-scale sweep ensures a balanced trade-off between stability and objective optimization.

\subsection{IQL Baseline Implementation}
For the ESR-based IQL baseline, we adopt the same network architecture as AETDICE to ensure a fair comparison. In particular, we use identical state encoders, symlog-transformed sinusoidal embeddings for the accumulated return inputs, and three-layer multilayer perceptrons with the same hidden dimensions.

We tune the expectile parameter $\tau$ over $\{0.8, 0.85, 0.9\}$ and the temperature parameter $\beta_{\text{IQL}}$ over $\{10, 12, 14, 16, 18\}$. IQL models are trained for $3\times10^{5}$ gradient steps using the Adam optimizer with the same learning rate and batch size as AETDICE.

\begin{table}[!htbp]
\centering
\caption{Hyperparameters for AETDICE on MO-PointMaze.}
\begin{tabular}{lc}
\toprule
$\pi_\theta$ learning rate & $3\times10^{-4}$ \\
$\nu_\psi$ learning rate & $3\times10^{-4}$ \\
$\mu$ learning rate & $3\times10^{-4}$ \\
Batch size & 256 \\
Hidden dim of $\nu_\psi$ and $\pi_\theta$ & 768 (1024 for MO-PointMaze-3obj) \\
n\_layer of $\nu_\psi$ and $\pi_\theta$ & 3 \\
Training steps & $10^{6}$ \\
Divergence $f$ & Pearson $\chi^2$ \\
$\beta$ & \shortstack{$\{10^{-6}, 5\times10^{-7}, 10^{-7}, 5\times10^{-8}, 10^{-8},$\\
         $5\times10^{-9}, 10^{-9}, 5\times10^{-10}, 10^{-10}\}$} \\
\bottomrule
\end{tabular}
\label{tab:aetdice_hparams}
\end{table}

\begin{table}[!htbp]
\centering
\caption{Hyperparameters for AETDICE on D4MORL~\cite{zhu2023scaling} (MO-Ant-v2, MO-Walker2d-v2, MO-HalfCheetah-v2).}
\begin{tabular}{lc}
\toprule
$\pi_\theta$ learning rate & $3\times10^{-4}$ \\
$\nu_\psi$ learning rate & $3\times10^{-4}$ \\
$\boldsymbol{\mu}$ learning rate & $3\times10^{-4}$ \\
Batch size & 512 \\
Hidden dim of $\nu_\psi$ and $\pi_\theta$ & 512 \\
n\_layer of $\nu_\psi$ and $\pi_\theta$ & 4 \\
Training steps & $10^{6}$ \\
Divergence $f$ & Pearson $\chi^2$ \\
$\beta$ & $\{10^{-4},\, 10^{-5},\, 10^{-6}\}$ \\
\bottomrule
\end{tabular}
\label{tab:aetdice_hparams_mujoco}
\end{table}

\section{Computational Resources}
All experiments were run on a single workstation with an Intel® Xeon® Gold 6330 CPU (256 GB RAM) and an NVIDIA RTX 3090 GPU. Training one AETDICE or IQL policy per task took approximately 15-20 minutes on average, with GPU memory consumption staying below 20 GB throughout training.

\section{Limitation and Impact statement}
\label{app:limitations}
\paragraph{Limitation}
While AETDICE enables offline optimization of a broader class of nonlinear MORL objectives, it inherits the fundamental challenges of offline MORL. First, as discussed in the ablation study (Appendix~\ref{app:ablation_beta}), offline MORL datasets are collected from policies trained under diverse preference weights, making trajectories optimal for one objective highly suboptimal for another. This requires smaller regularization to permit larger distribution shift, making offline optimization more demanding than in the single-objective setting.
Second, the augmented state space introduces additional scalability challenges. Incorporating $\mathbf{R}^{\mathrm{acc}}_t$ into the state increases the dimensionality by $m$ (the number of objectives), and the data must provide sufficient coverage over this augmented space. As the number of objectives grows, the coverage requirement becomes increasingly difficult to satisfy, potentially limiting the applicability of AETDICE to problems with a large number of objectives.

\paragraph{Impact statement} AETDICE advances offline MORL by enabling optimization of nonlinear objectives that capture fairness and balanced performance across multiple criteria. This is relevant to real-world equitable decision-making, where policies must balance competing objectives both within and across episodes. As a methodological contribution, we do not anticipate direct negative societal impacts. However, as with any offline RL method, the quality of learned policies depends on the data, and biases present in the dataset may propagate to the optimized policy.

\newpage

\end{document}